\documentclass{article}
\usepackage{caption}
\usepackage{microtype}
\usepackage{subfigure}
\usepackage{booktabs} 
\usepackage{hyperref}
\usepackage{listings}
\usepackage{xcolor}
\definecolor{codegreen}{rgb}{0,0.6,0}
\definecolor{codegray}{rgb}{0.5,0.5,0.5}
\definecolor{codepurple}{rgb}{0.58,0,0.82}
\definecolor{backcolour}{rgb}{0.95,0.95,0.92}

\lstdefinestyle{mystyle}{
    backgroundcolor=\color{backcolour},   
    commentstyle=\color{codegreen},
    keywordstyle=\color{magenta},
    numberstyle=\tiny\color{codegray},
    stringstyle=\color{codepurple},
    basicstyle=\ttfamily\footnotesize,
    breakatwhitespace=false,
    breaklines=true,         
    captionpos=b,
    keepspaces=true,
    numbers=left,
    numbersep=5pt,
    showspaces=false,
    showstringspaces=false,
    showtabs=false,
    tabsize=1
}
 
\usepackage{graphicx}
\usepackage[accepted]{icml2024}

\usepackage{amsmath}
\usepackage{amssymb}
\usepackage{mathtools}
\usepackage{amsthm}

\usepackage[capitalize,noabbrev]{cleveref}

\usepackage{multirow}
\usepackage{float}
\usepackage{makecell}

\theoremstyle{plain}
\newtheorem{theorem}{Theorem}[section]
\newtheorem{proposition}[theorem]{Proposition}

\theoremstyle{definition}
\newtheorem{definition}[theorem]{Definition}
\newtheorem{assumption}[theorem]{Assumption}
\theoremstyle{remark}

\icmltitlerunning{Guidance with Spherical Gaussian Constraint for Conditional Diffusion}

\begin{document}

\twocolumn[
\icmltitle{Guidance with Spherical Gaussian Constraint for Conditional Diffusion}

\icmlsetsymbol{equal}{*}
\begin{icmlauthorlist}
\icmlauthor{Lingxiao Yang}{shanghaitech,MoE}
\icmlauthor{Shutong Ding}{shanghaitech,MoE}
\icmlauthor{Yifan Cai}{shanghaitech,MoE}
\icmlauthor{Jingyi Yu}{shanghaitech,MoE}
\icmlauthor{Jingya Wang}{shanghaitech,MoE}
\icmlauthor{Ye Shi}{shanghaitech,MoE,equal}
\end{icmlauthorlist}

\icmlaffiliation{shanghaitech}{ShanghaiTech University}
\icmlaffiliation{MoE}{MoE Key Laboratory of Intelligent Perception and Human Machine Collaboration}

\icmlcorrespondingauthor{Ye Shi}{shiye@shanghaitech.edu.cn}

\icmlkeywords{Machine Learning, ICML}

\vskip 0.3in
]

\printAffiliationsAndNotice{} 

\begin{figure*}[h]
  \centering
  \includegraphics[width=1\linewidth]{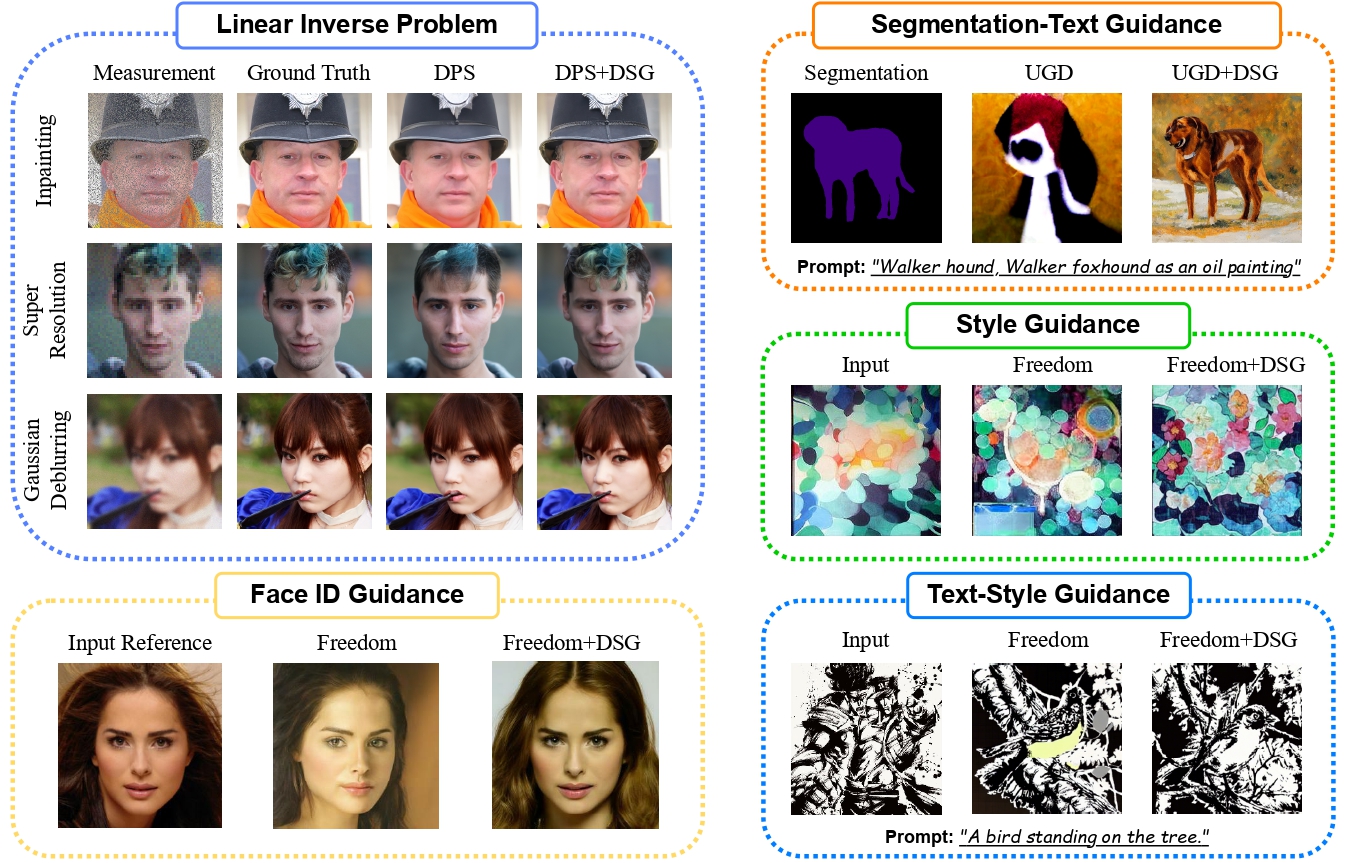} 
   \vspace{-0.4cm}
    \caption{DSG mitigates manifold deviation problem by introducing Spherical Gaussian constraint without relying on the linear manifold assumption. Simultaneously, DSG enables the use of larger step sizes, significantly reducing inference time while enhancing sample quality. The integration of DSG into existing training-free CDMs only incurs a few additional lines of code. Here we briefly show the performance of integrating DSG into recent CDMs, such as DPS \cite{dps} for Inpainting, Super Resolution, Gaussian Deblurring, UGD \cite{Universal} for Segmentation-text Guidance, and Freedom \cite{freedom} for Style Guidance tasks. DSG imposes nearly negligible computational overhead while delivering substantial performance enhancements.}
\label{fig1} 
\end{figure*}

\begin{abstract}
Recent advances in diffusion models attempt to handle conditional generative tasks by utilizing a differentiable loss function for guidance without the need for additional training. While these methods achieved certain success, they often compromise on sample quality and require small guidance step sizes, leading to longer sampling processes. This paper reveals that the fundamental issue lies in the manifold deviation during the sampling process when loss guidance is employed. We theoretically show the existence of manifold deviation by establishing a certain lower bound for the estimation error of the loss guidance. To mitigate this problem, we propose Diffusion with Spherical Gaussian constraint (DSG), drawing inspiration from the concentration phenomenon in high-dimensional Gaussian distributions. DSG effectively constrains the guidance step within the intermediate data manifold through optimization and enables the use of larger guidance steps. Furthermore, we present a closed-form solution for DSG denoising with the Spherical Gaussian constraint. Notably, DSG can seamlessly integrate as a plugin module within existing training-free conditional diffusion methods. Implementing DSG merely involves a few lines of additional code with almost no extra computational overhead, yet it leads to significant performance improvements. Comprehensive experimental results in various conditional generation tasks validate the superiority and adaptability of DSG in terms of both sample quality and time efficiency.
\end{abstract}


\section{Introduction}
\label{sect:intro}

The past few years have witnessed great successes of diffusion models \cite{sohl2015deep, ddpm,ncsn,sde}, especially conditional diffusion models (CDMs) \cite{classifier,classifierfree, stable, control}, for their powerful expressive and re-editing abilities in generative tasks such as image generation \cite{stable,iddpm,incsn,song2021maximum}, image inpainting \cite{mcg,lugmayr2022repaint,dps}, super-resolution \cite{daniels2021score,saharia2022image}, image editing \cite{choi2021ilvr,meng2021sdedit}, and human motion generation \cite{tevet2022human,zhang2022motiondiffuse}. 

Generally, there are two principal techniques in conditional diffusion models, known as classifier-guided \cite{classifier} and classifier-free \cite{classifierfree} diffusion models respectively. However, there still exist challenges in learning cost and model generality since these two methods both need extra training and data to endow the diffusion model with conditional generation ability. Hence, recent advances \cite{mcg, dps, DiffPIR, Universal, freedom, manifold} in conditional diffusion models develop training-free methods, which only use the off-the-shelf loss guidance during the denoising process. While these training-free methods avoid extra training, they may sacrifice the realism and quality of the generated samples. This weakness is probably due to the deviation between the intermediate data sample $x_t$ from the intermediate data manifold during the denoising process. Due to this issue, existing methods \cite{dps, freedom, Universal} usually had to adopt small loss-guided step sizes but this would further increase the number of sampling steps and make their algorithms slow. 

In this paper, we reveal the occurrence of manifold deviation during the diffusion sampling process by establishing a certain lower bound for the estimation error of the loss guidance. To address the 
manifold deviation issue, we propose \textbf{D}iffusion with \textbf{S}pherical \textbf{G}aussian constraint (DSG\footnote{Code is available at \url{https://github.com/LingxiaoYang2023/DSG2024}.}), a novel training-free method for the conditional diffusion model. The principle idea of DSG is to restrict the guidance step within the intermediate data manifold via the Spherical Gaussian constraint. Concretely, the Spherical Gaussian constraint is a spherical surface determined by the intermediate data manifold, which is the high-confidence region of the unconditional diffusion step. Then, we formulate the calculation of guidance as an optimization problem with the Spherical Gaussian constraint and the guided-loss objective. In that case, the manifold deviation can be avoided and a sufficiently large guidance step size can also be adopted. Besides, we find that this optimization problem can be effectively solved with a closed-form solution. This means DSG imposes almost no extra computational costs in each diffusion step. Hence, our model achieves better performances in both time efficiency and sample quality compared with previous training-free diffusion models. To summarize, our contribution is summarized as follows:

    \textbf{1) Reveal the manifold deviation issue.} We illustrate the intermediate data manifold deviation problem that exists widely in previous training-free conditional diffusion models by showing the certain lower bound for the estimation error of the loss guidance, as induced by the Jensen gap. 

    \textbf{2) Spherical-Gaussian-constrained guidance.} We introduce the Diffusion with Spherical Gaussian Constraint (DSG) method, a plug-and-play module for training-free conditional diffusion models. DSG is designed based on the high-dimensional Gaussian distribution concentration phenomenon, effectively mitigating the manifold deviation issue. 

    \textbf{3) A few lines of code to boost performance.} 
    We provide a closed-form solution for the DSG denoising process, simplifying the integration of DSG into existing training-free conditional diffusion models to just a few additional lines of code. DSG incurs nearly negligible computational overhead while delivering substantial performance enhancements.
    
    \textbf{4) Applicable for various tasks}. We conduct experiments on various conditional generation tasks, involving Inpainting, Super Resolution, Gaussian Deblurring, Text-Segmentation Guidance, Style Guidance, Text-Style Guidance, and FaceID Guidance. These results robustly confirm the superiority and adaptability of our method in terms of both sample quality and time efficiency. 

\section{Related Work}
\label{sect:related}
In this section, we briefly review the literature on training-required and training-free conditional diffusion models (CDMs). 

\textbf{Training-required CDMs.} Generally, training-required CDMs are divided into two principal branches. One of them is the classifier-guided diffusion model \cite{classifier}, which trains a time-dependant classifier, based on an off-the-shelf diffusion model, to approximate an unbiased posterior probability $p(y|x_t)$. Another branch is the classifier-free diffusion model \cite{classifierfree} such as the well-known Stable Diffusion \cite{stable} and variants of it \cite{control, mou2023t2i}. Different from the classifier-guided model, The key idea of it is to utilize the data pairs and train a conditional denoiser directly. While these models can ensure the realism and quality of the generated samples with the given condition, it is obvious that the extra cost of data-pair collection and training time training is unavoidable. This makes it hard to deploy them in some real-world scenarios without extra data pairs or sufficient computation resources.

\textbf{Training-free CDMs.} To overcome the issue, recent research works attempt to handle conditional generation in a training-free manner. Generally, instead of training the time-dependent classifier, they utilize the off-the-shelf classifier guidance to approximate $\nabla_{x_t} \log(y|x_t)$. Specifically, MCG \cite{mcg} approximated the guidance with the aid of Tweedie's formula to solve linear inverse problems, and then this idea is extended to general conditional generation tasks by DPS\cite{dps}, Freedom \cite{freedom}, UGD \cite{Universal}. Besides, LGD-MC \cite{Lossguided} attempted to use multiple samples from an inaccurate Gaussian distribution to reduce the estimation bias but it still suffers from the computation cost for Monte Carlo simulation. Other works like UGD, DiffPIR \cite{Universal, DiffPIR} performed the guidance step on the clean data sample (i.e., $x_0$) and then projected the updated clean data sample to the corresponding intermediate data manifold (i.e., $x_t$). Very recently, a concurrent work called MPGD \cite{manifold} further studied this idea with an extra auto-encoder via enforcing the guidance step within the tangent space of the clean data manifold. While it makes some improvements, it depends on the strong linear assumption of the data manifold, which does not hold in most real-world applications. Additionally, MPGD heavily relies on pre-trained autoencoders for manifold projections. While it performs well in image generation, it cannot generalize to scenarios where pre-trained autoencoders are not available. In contrast, the proposed DSG method can handle the manifold deviation problem properly without such strong assumptions and improve the performance on both sample quality and time efficiency.

\section{Preliminaries}

\subsection{Diffusion Models}
Diffusion models \cite{ddpm,ncsn,sde} are a class of likelihood-based generative models that outperform other generative models(e.g. GAN \cite{gan}, VAE \cite{vae}) in various tasks. It has predefined forward processes, where Gaussian noise is incrementally added to clean data $x_0$ until it evolves into pure noise $x_T \in \mathcal{N}(0, I)$. In the formulation of DDPM \cite{ddpm}, the posterior distribution of any $x_t, t \in [0,T]$ given  $x_0$ is defined as:
\begin{equation}
    q(x_t | x_0) = \mathcal{N}(x_t; \sqrt{\alpha_t}x_0 ,(1-\alpha_t)I).
\end{equation}
Here $\{\alpha_0, \cdots, \alpha_T\}$ is the predefined noise schedule. The objective is to train a denoiser, denoted as $\epsilon_\theta (x_t,t)$, to approximate Gaussian noise $\epsilon_t$ at each time step $t$. This is achieved by minimizing the loss function $L_\text{simple}$, which is equivalent to matching the denoising scores $\nabla_{x_t} \log p_t(x_t)$ \cite{ddpm}.
\begin{equation}
    L_\text{simple} = \mathbb{E}_{t,x_0,\epsilon_t}[||\epsilon_t - \epsilon_\theta (x_t,t)||^2].
\end{equation}
During the sampling stage, the joint distribution $p(x_{0:T})$ in the reverse process can be defined as: 
\begin{equation}
    p_\theta (x_{0:T})=p(x_T) \prod_{t=1}^{T} p_\theta(x_{t-1} | x_t).
\end{equation}
\begin{equation}
    \label{eq:sample}
    p_\theta (x_{t-1} | x_t)=\mathcal{N}(x_{t-1}; \mu_\theta (x_t,t), \Sigma_\theta (x_t, t)).
\end{equation}
In this work, we employ the update rule of DDIM \cite{ddim} for calculating $\mu_\theta (x_t,t)$ and $ \Sigma_\theta (x_t, t))$:
\begin{equation}
\label{eq:ddim}
\begin{split}
    x_{t-1}=\sqrt{\alpha_{t-1}} \underbrace{\left(\frac{x_{t}-\sqrt{1-\alpha_{t}} \epsilon_{\theta}^{(t)}\left(x_{t}\right)}{\sqrt{\alpha_{t}}}\right)}_{\text {" predicted } x_{0} \text { " }}+ \\ \underbrace{\sqrt{1-\alpha_{t-1}-\sigma_{t}^{2}} \cdot \epsilon_{\theta}^{(t)}\left(x_{t}\right)}_{\text {"direction pointing to } x_{t} \text { " }}+\underbrace{\sigma_{t} \epsilon_{t}}_{\text {random noise }},
\end{split}
\end{equation}
where $\sigma_t = \eta \sqrt{(1- \alpha_{t-1} / (1- \alpha_t))}\sqrt{1- \alpha_t / \alpha_{t-1}}$. When $\eta=1$, the generative process becomes DDPM, while $\eta=0$ corresponds to deterministic sampling.

\subsection{Training-Free Conditional Diffusion Methods} 

Classifier guidance \cite{classifier} is the first work that utilizes the pre-trained diffusion model in conditional generative tasks. Specifically, considering the Bayes rule $p(x|y)=p(y|x)p(x)/p(y)$, it introduces the given condition with an additional likelihood term $p(x_t | y)$:
\begin{equation}
    \nabla_{x_t} \log p(x_t | y) = \nabla_{x_t} \log p(x_t) + \nabla_{x_t} \log p(y | x_t), 
\end{equation}
where $y$ represents the condition or measurement. Recently, many works, known as training-free methods, attempt to employ the pre-trained diffusion models for the conditional generation without re-training or fine-tuning the diffusion prior. Without training the time-dependent classifier to estimate $p(y | x_t)$, these current training-free guided diffusion models just need a pre-trained diffusion prior $D_\theta$ and a differentiable loss function $L(x_0,y)$ which is defined on the support of $x_0$ using an off-the-shelf neural network. They use Tweedie’s formula to calculate $\nabla_{x_t} \log p(y | x_t)$ by estimating $\hat{x}_0$ based on $x_t$:
\begin{equation}
    \hat{x}_0(x_t) \approx \mathbb{E}[x_0 | x_t] = (x_t + b_t^2 \nabla_{x_t} \log p(x_t)) / a_t,
\end{equation}
\begin{equation}
    \begin{split}
        \nabla_{x_t} \log p(y | x_t) &\approx  \nabla_{x_t} \log p(y | \hat{x}_0(x_t)), \\
                                     &= \gamma \nabla_{x_t} L(\hat{x}_0(x_t),y)
    \end{split}                
\end{equation}
where $x_t \sim \mathcal{N}(a_t x_0, b_t^2 I)$ and $\gamma$ is the step size. In the formulation of DDPM \cite{ddpm}, $a_t = \sqrt{\alpha_t}$, $b_t^2 = \sqrt{1-\alpha_t}$. Then they use the estimated likelihood of $L(x_0,y)$ for additional correction step:
\begin{equation}
    x_{t-1} = \underbrace{DDIM(x_t, \epsilon_\theta(x_t,t))}_{\text{sampling step}} - \underbrace{\gamma \nabla_{x_t} L(\hat{x}_0(x_t),y)}_{\text{correction step}}.
\end{equation}

\subsection{Low-dimensional Manifold Assumption}
Suppose \textit{clean data manifold} $\mathcal{M}_0$ is the set consisting of all data points. Typically, the data are assumed to lie in a low-dimensional space rather than a whole ambient space:
\begin{assumption} 
(Low-dimensional Manifold Assumption). The data manifold $\mathcal{M}_0$ lies in the $k$-dimensional subspace $\mathbb{R}^k$ with $k \ll n$. 
\end{assumption}
Given this assumption, recent work \cite{mcg} has shown that the set of noisy data $x_t$, a.k.a. \textit{intermediate data manifold} in this paper, is concentrated on a ($n-k$)-dimensional manifold $\mathcal{M}_t: \{x \in \mathbb{R}^n | d(x,\sqrt{\alpha_t}\mathcal{M}) = \sqrt{(1-\alpha_t)(n-k)}\}$. 

\section{Method}

In this section, we will first present the limitations of previous training-free CDMs, including the strong manifold assumption and estimation error of the loss guidance. After that, our proposed algorithm, DSG, will be proposed to effectively address and resolve these issues by applying guidance with the spherical Gaussian constraint.

\subsection{Manifold Deviation by Loss Guidance}
\label{sect:deviation}

Although previous training-free work has achieved great success in various image fields due to its plug-and-play characteristics, they will sacrifice the quality of generated samples. There are two main issues for this phenomenon: linear manifold assumption about $\mathcal{M}_0$ and the estimation error of loss guidance induced by the Jensen gap, which leads to the manifold deviation problem.

\textbf{Linear manifold assumption.} Except for the low-dimensional manifold assumption, some of the previous training-free conditional diffusion methods \cite{mcg, manifold} further require a linear subspace assumption for $\mathcal{M}_0$. MCG \cite{mcg} employed this linear subspace assumption to establish the estimated gradient $\nabla_{x_t} L(\hat{x}_0(x_t),y)$. A concurrent work, MPGD \cite{manifold}, projected the gradient of the loss on $\hat{x}_0$ to the tangent space of $M_0$ to keep the sample located on the intermediate manifold $\mathcal{M}_t$ for any $t \in [0, T]$. But this projection is also based on the linear manifold assumption. However, the linear manifold assumption is quite strong, so in practice, both MCG and MPGD will inevitably introduce errors in estimating $p(y | x_t)$, leading to deviations of generated samples from the clean data manifold $\mathcal{M}_0$.

\textbf{Jensen gap.} As mentioned in DPS \cite{dps}, the estimation error of loss guidance results in previous works results from the approximation of  $\mathbb{E}_{x_0|x_t}\left[L(x_0)\right] \approx L\left(\mathbb{E}_{x_0|x_t}\left[x_0\right]\right) = L(\hat{x}_0(x_t))$. Actually, the gap between $\mathbb{E}_{x_0|x_t}\left[L(x_0)\right]$ and $L(\hat{x}_0(x_t))$ is related to Jensen's inequality, which is known as the Jensen gap.
\begin{definition}
(Jensen gap). Let $x$ be a random variable with distribution $p(x)$. For function $f$ that is convex or non-convex, the Jensen gap is defined as:
\begin{equation}
    \mathcal{J}(f,x \sim p(x)) = \mathbb{E}[f(x)] - f(\mathbb{E}[x]),
\end{equation}
where the expectation is taken over $p(x)$.
\end{definition}
Although DPS~\cite{dps} provides an upper bound for the Jensen gap, it depends on the assumption on the finite derivatives of guided loss, which is obviously unavailable at least in linear inverse problems. In fact, the lower bound of this gap is very large in high-dimensional generative tasks, as shown in Proposition \ref{prop:lower}, and it further leads to the serious manifold deviation problem.
\begin{proposition}
\label{prop:lower}
(\textit{Lower bound for Jensen gap}). For the $\beta$-strongly convex function $f$ and the random variable $x\in \mathbb{R}^n \sim \mathcal{N}(\mu, \Sigma)$, we can have the lower bound for the Jensen Gap:
\[
    \mathcal{J} \ge \frac{1}{2}\beta \sum_{i=1}^n \lambda_i.
\]
where $\Sigma = P^T\Lambda P$ via spectral decomposition and $\lambda_1,\cdots,\lambda_n$ are positive diagonal elements in $\Lambda$.
\end{proposition}
Please refer to the Appendix for our proof. In specific, the function $f$ is the off-the-shelf guided loss $L$ and the random variable $x$ is $x_0\left|x_t\right.$. Notably, here we utilize the assumption on $x_0|x_t$ mentioned in LGD-MC \cite{Lossguided} and extend it to general Gaussian distribution for more accurate estimation. Besides, the assumption on strong convexity of $L$ is also rational. On one hand, $L$ is usually a quadratic function in linear inverse problems, which is $\beta$-strongly convex. On the other hand, in practical scenarios, $L$ tends to possess at least local $\beta$-strong convexity. Considering the probability of Gaussian distribution can be ignored when $x$ is far from $\hat{x}_0$, the lower bound could still hold in general cases. Hence, the Jensen gap increases linearly with the number of sample dimensions, and it will be very large in high-dimensional generative tasks like image generation according to Proposition \ref{prop:lower}. In that case, such a large gap leads to the deviation of $x_{t-1}$ from the intermediate data manifold $\mathcal{M}_{t-1}$, and thus causes the loss of image structural information. 

\subsection{Diffusion with Spherical Gaussian Constraint}

\begin{figure}[t]
  \centering
  \includegraphics[width=1\linewidth]{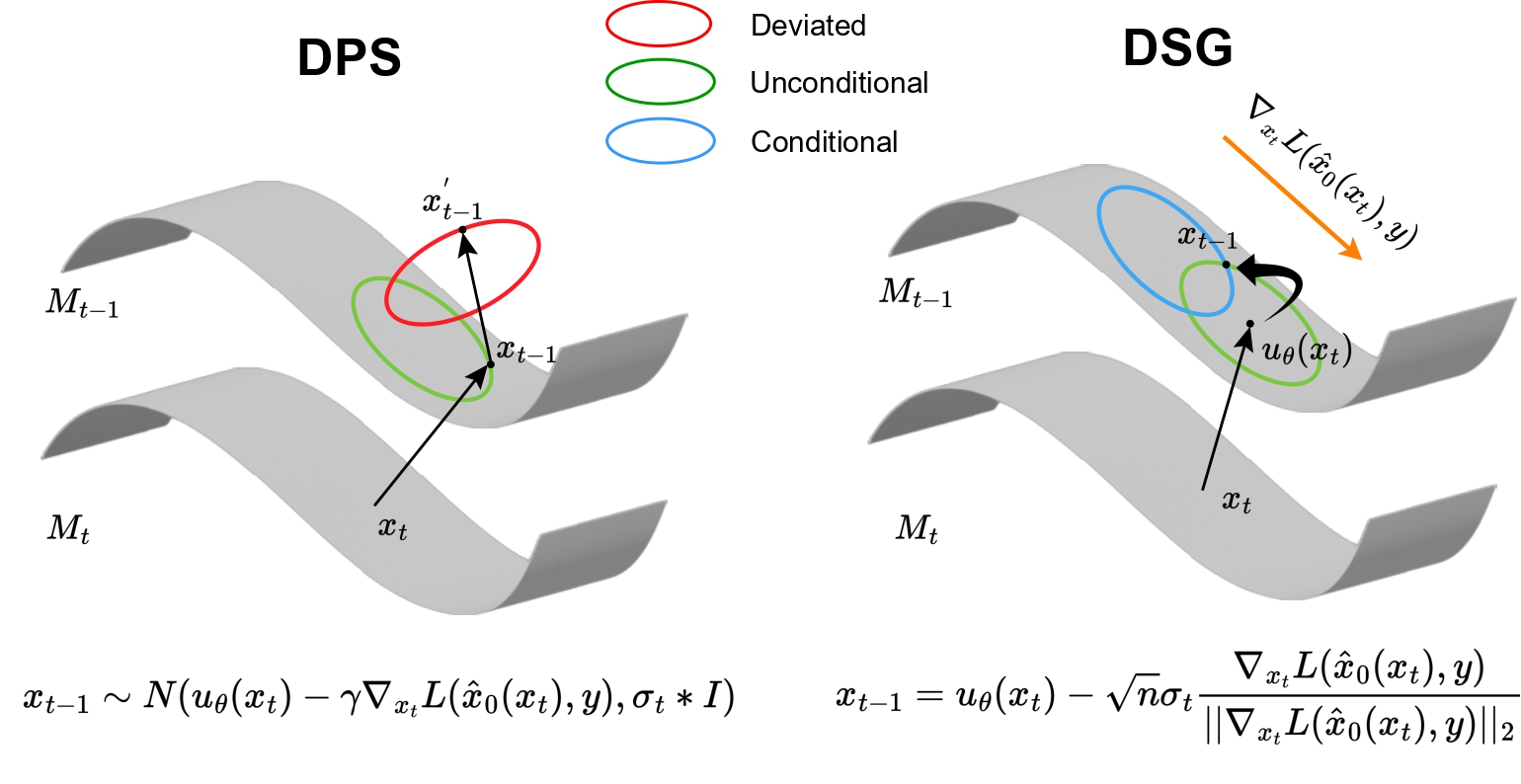} 
   \vspace{-0.8cm}
    \caption{Illustration of the manifold deviation problem in DPS (left) and a schematic overview of how our DSG restricts the guidance within the manifold (right). The red circular ring represents the concentration region of samples under deviated conditional guidance, the blue circular ring indicates the concentration region of samples under accurate conditional guidance, and the green circular ring represents the concentration region of samples without condition.}
  \vspace{-0.7cm}
\label{fig:method} 
\end{figure}
According to our theoretical analysis of previous methods in Section \ref{sect:deviation}, both the Jensen gap and the linear manifold assumption will introduce additional errors, leading to a substantial decline in the authenticity of generated samples in comparison to unconditional generation.

Taking an alternative perspective, considering the inevitable error in estimating the conditional bias in every denoising step, why not start from the pre-existing unconditional intermediate manifold $\mathcal{M}_t$ and subsequently find the point closest to the conditional sampling?

To this end, we propose diffusion with spherical Gaussian constraint (DSG), an optimization method that performs guidance within high confidence intervals of unconditional intermediate manifold $\mathcal{M}_t$:
\begin{equation}
    \label{eq:dsg1}
    \begin{split}
        \mathop{\arg\min}\limits_{x'} &  \left[\nabla_{x_t} L(\hat{x}_0(x_t),y)\right]^T (x'-x_t)  \\
        \text{s.t.}& \;x' \in CI_{1-\delta}
    \end{split}
\end{equation}
where $CI_{1-\delta}$ represents the ($1-\delta$) confidence intervals for Gaussian distribution in Equation (\ref{eq:sample}).

In this optimization problem, the objective encourages sampling in the direction of gradient descent, while the constraint enforces sampling within high-confidence intervals for a Gaussian distribution.

Nevertheless, when high-confidence intervals encompass substantial regions in n-dimensional space, the optimization problem may appear to be challenging and infeasible. Fortunately, in the case of high-dimensional isotropic Gaussian distribution, where high-confidence intervals are concentrated on a hypersphere, we can simplify the constraint by approximating it with this hypersphere, referred to as the spherical Gaussian constraint.
\begin{proposition}
\label{pro:concentration}
(Concentration of high-dimensional isotropy Gaussian distribution). For any n-dimensional isotropy Gaussian distribution $x \sim \mathcal{N}(\mu,\sigma^2 I)$:
\begin{equation}
    \begin{split}
        P(||x-\mu||_2^2 \geq  x_{lower}=n\sigma^2+2n\sigma^2 (\sqrt{\epsilon}+\epsilon)) \leq e^{-n\epsilon}, \nonumber \\ 
        P(||x-\mu||_2^2 \leq x_{upper}=n\sigma^2 - 2n\sigma^2\sqrt{\epsilon}) \leq e^{-n\epsilon}, \nonumber\\
    \end{split}
\end{equation}
When $n$ is sufficiently large, it is close to the uniform distribution on the hypersphere of radius $\sqrt{n}\sigma$.
\end{proposition}
\begin{algorithm}[tb]
   \caption{Diffusion with Spherical Gaussian constraint}
   \label{algo:dsg}
\begin{algorithmic}

   \STATE {\bfseries Input:} pure noise $x_T \sim N(0,I)$, guidance interval $i$, guidance rate $g_r$
   \FOR{$t=T$ {\bfseries to} $1$}
       \STATE $ \epsilon_t \sim N(0, I)$  
        \STATE $ \hat{x}_0(x_t) = (x_t - \sqrt{1-\alpha_t} 
     \epsilon_\theta(x_t,t)) / \sqrt{\alpha_t}$
        \STATE $ \mu_{\theta}(x_t)=\sqrt{\alpha_{t-1}} \hat{x}_0(x_t) + \sqrt{1-\alpha_{t-1}-\sigma_{t}^{2}} \cdot \epsilon_{\theta}(\boldsymbol{x}_{t},t)$

     \STATE \;\textcolor[rgb]{0.5,0.5,0.5}{//\,Guidance with Spherical Gaussian Constraint }

     \STATE $d^{*} = - \sqrt{n}\sigma_t \cdot  \frac{\nabla_{x_t}L(\hat{x}_0(x_t),y)}{||\nabla_{x_t}L(\hat{x}_0(x_t),y)||}$

    \STATE $d^{sample} = \sigma_t \epsilon_t$

    \STATE $d_m = d^{sample} + g_r(d^*-d^{sample})$
    \STATE $x_{t-1} = \mu_\theta (x_t) + r \frac{d_m}{||d_m||} $
     
   \ENDFOR
   \STATE \textbf{Return} $x_0$
   
\end{algorithmic}
\end{algorithm}

Please refer to Appendix \ref{app:gaussian} for further proof. According to Proposition \ref{pro:concentration}, since the probability of both $P(||x-\mu||_2^2 \geq x_{lower})$ and $P(||x-\mu||_2^2 \leq x_{upper})$ decrease sharply with the increase of $\epsilon$, the posterior Gaussian distribution in Equation \ref{eq:ddim} can be approximated to hypersphere $S^n_{\mu,r}=S^n_{\mu_\theta(x_t),\sqrt{n}\sigma}=\{x \in \mathbb{R}^n: ||x-\mu_\theta(x_t)||_2^2= r^2=n \sigma_t^2 \}$. Hence, the optimization problem can be approximated as:
\begin{equation}
    \label{eq:dsg2}
    \begin{split}
        \mathop{\arg\min}\limits_{x'} & \left[\nabla_{x_t} L(\hat{x}_0(x_t),y)\right]^T (x'-x_t)  \\
        \text{s.t.}& \;x' \in S^n_{\mu_\theta(x_t),\sqrt{n}\sigma_t}
    \end{split}
\end{equation}

\begin{figure*}[htbp]
  \centering
  \vspace{-0.1cm}
  \includegraphics[width=0.8\linewidth]{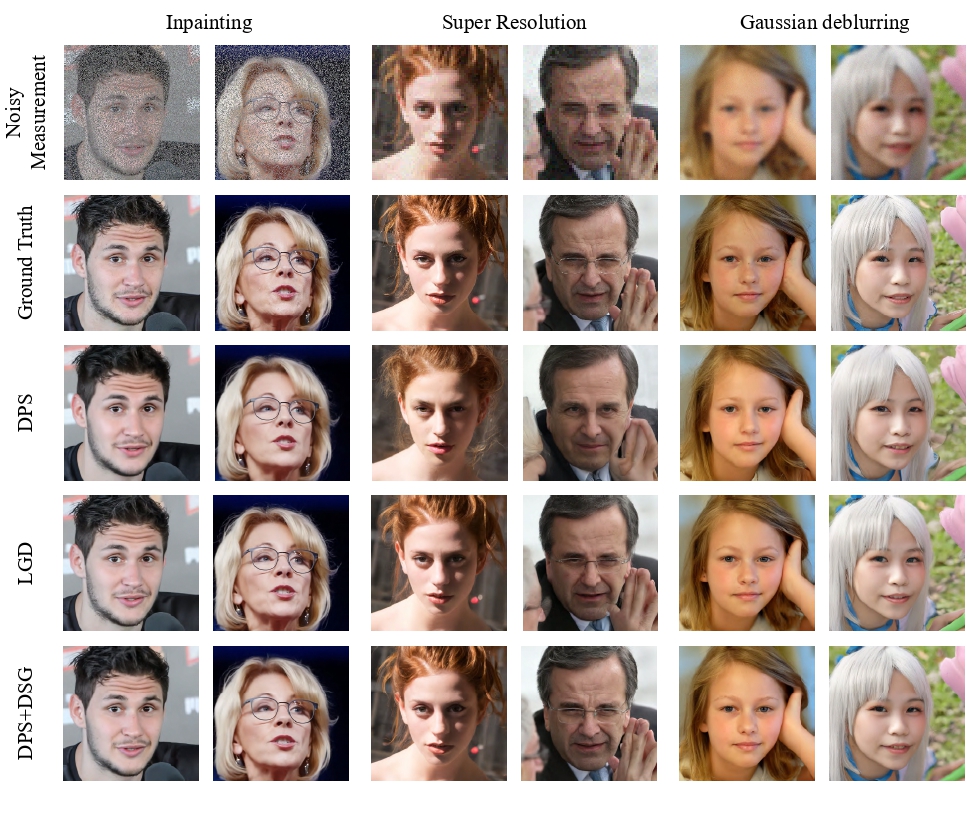}
  \vspace{-0.9cm}
  \caption{Result in solving three linear inverse problems (Inpainting, Super-resolution, Gaussian deblurring) using DSG using the pre-trained FFHQ diffusion model. }
  \vspace{-0.5cm}
\label{fig:lin} 
\end{figure*}

While this optimization problem is inherently non-convex owing to the spherical Gaussian constraint, we can obtain a closed-form solution: 
\begin{equation}\label{closed-form}
x^*_{t-1} = \mu_\theta(x_t) - \sqrt{n}\sigma_t \frac{\nabla_{x_t} L(\hat{x}_0(x_t),y)}{||\nabla_{x_t} L(\hat{x}_0(x_t),y)||_2}. 
\end{equation}
Detailed derivation is provided in the Appendix \ref{app:solution}.

\textbf{Remark 1.} It's worth noting that our DSG can be a plugin module to existing training-free CDMs, such as DPS \cite{dps}, UGD \cite{Universal}, Freedom \cite{freedom}, etc, while incurring almost no extra computational overhead. Notably, this method only requires modifying a few lines of code to generate more realistic images with accelerated sampling speed. 
Using DPS \cite{dps} as an example, we highlight the key difference of code between DPS and our DSG as follows: 

\begin{lstlisting}[language=Python]
diff=measure-self.forward(x_0_hat)
dist=norm(diff)
grad=autograd.grad(
         outputs=distance, inputs=x_t)[0]
# DPS: x_t=x_t_mean+sigma_t*eps_t-         grad*self.gamma
# Ours:
norm=norm(grad, dim=[1, 2, 3])
x_t=x_t_mean-sqrt(n)*sigma_t*grad/norm
\end{lstlisting}

\textbf{Remark 2.}
Another intuitive explanation for the optimal solution obtained is that we directly apply the gradient descent to $\mu_\theta (x_t)$ instead of sampling point $x_t$. While the standard deviation $\sigma_t$ of DDIM generally decreases as time $t$ gets smaller, our method can be seen as adaptive gradient descent which decreases fast in the early stage and then coverage in the final stage. In practice, we can use extremely large step sizes (exceed 400$\times$ when we use $\eta=1$ and $t=T$) compared to current training-free methods, thus achieving better alignment. Therefore, our method is robust to smaller DDIM steps while the performance of DPS quickly degrades, which will be evaluated in Sec \ref{sec:ablation}.

In practice, when employing DSG to enhance alignment and authenticity, it may sacrifice sample diversity. Therefore, we use the weighted direction of the optimized gradient and unconditional sampling like the Classifier-free \cite{classifierfree} manner to enhance diversity:
\begin{equation}
    d_m = d^{sample} + g_r(d^*-d^{sample}).
    \label{eq:mix}
\end{equation}
\begin{equation}
    x_{t-1} = \mu_\theta (x_t) + r \frac{d_m}{||d_m||}.
\end{equation}
Here $d^{sample}=\sigma_t \epsilon_t$ represents the unconditional sampling direction, $d^*=- \sqrt{n}\sigma_t \cdot  \frac{\nabla_{x_t}L(\hat{x}_0(x_t),y)}{||\nabla_{x_t}L(\hat{x}_0(x_t),y)||}$ represents the steepest gradient descent direction, and $d_m$ represents the weighted direction which will be scaled to satisfy the Spherical Gaussian constraint. The detailed procedure is shown in Algorithm \ref{algo:dsg}.

\section{Experiments}

In this section, we evaluate our methods in various tasks, including three linear inverse problems ( Inpainting, Super-resolution, Gaussian deblurring), Style Guidance, Text-Segmentation Guidance, FaceID Guidance, and Text-Style Guidance. We demonstrate that our method can plug and play easily into various training-free methods including DPS \cite{dps}, Freedom \cite{freedom}, and UGD \cite{Universal}, and improve their performance significantly.

\subsection{Implementation Details}

\textbf{Linear Inverse Problems} For linear inverse problem $y = Ax + \epsilon$ including Inpainting, Super-resolution, and Gaussian deblurring, we evaluate our method in 1k images of FFHQ256$\times$256 \cite{ffhq} and Imagenet256$\times$256 validation dataset \cite{deng2009imagenet} using pre-trained diffusion models taken from \cite{dps, classifier}. For comparison, we choose MCG \cite{mcg}, PnP-ADMM \cite{chan2016plug}, Score-SDE \cite{sde}, DDRM \cite{ddrm}, DPS\cite{dps} and LGD \cite{Lossguided} as the baseline. We generate noisy measurements by introducing Gaussian noise $\epsilon \sim \mathcal{N}(0,0.05)$. The noisy measurements are then acquired through various forward models $A(\cdot)$ customized for specific tasks: (i) In the context of image Inpainting, we randomly mask approximately 92\% of the total pixels within the RGB channel. (ii) For Super-resolution, we employ bicubic downsampling to achieve a 4x reduction in resolution. (iii) For Gaussian deblurring, we apply a 61x61 kernel size Gaussian blur with a standard deviation of 3.0. The loss guidance can be expressed as:
\begin{equation}
    \text{Loss}(x_0,y) = ||A\hat{x}_0(x_t)-y||_2^2,
\end{equation}
where $y$ represents the noisy measurement and $x_0$ donates the image we aim to reconstruct.

\begin{table}[h]
    \vspace{-0.3cm}
    \caption{Quantitative evaluation of the FFHQ dataset in Linear Inverse Problem }
    \label{table:comparison}
    \centering

    \resizebox{1\linewidth}{!}{
    \begin{tabular}{c |  cc | cc | cc | }
        \toprule[1.2pt]
          \multirow{2}{*}{Methods}  &  \multicolumn{2}{c}{Inpainting} \vline & \multicolumn{2}{c}{Super resolution} \vline & \multicolumn{2}{c}{Gaussian deblurring} \vline \\
        \rule{0pt}{8pt}

             & LPIPS↓ & FID↓ &  LPIPS↓ & FID↓ & LPIPS↓ & FID↓ \\
        \toprule[1.2pt]
       DDRM \cite{ddrm} & 0.665 & 114.9  & 0.339 & 59.57 & 0.427 & 63.02\\ 
       MCG \cite{mcg} & 0.414 & 39.19 & 0.637 & 144.5 &  0.550 & 95.04\\
       PnP-ADMM \cite{chan2016plug}  & 0.677 & 114.7 & 0.433 & 97.27 & 0.519 & 100.6\\ 
       \makecell{Score-SDE \cite{sde} \\ (ILVR \cite{choi2021ilvr})} & 0.659 &  127.1 & 0.701 & 170.7 & 0.667 & 120.3\\ 
       DPS \cite{dps} & 0.212 & 21.19 & 0.214 & 39.35 & 0.257 & 44.05 \\
       LGD \cite{Lossguided} & 0.159 & 28.21 & 0.231 & 34.44 & 0.229 & 32.57 \\
        \toprule[1.2pt]            

        DPS+DSG(Ours) &   \textbf{0.115} & \textbf{15.77}  & \textbf{0.211} & \textbf{30.30} &  \textbf{0.208} & \textbf{28.22} \\
  
        \bottomrule[1.2pt]
    \end{tabular}
    }
    \label{tb:comp_tedigan}
\end{table}
\begin{table*}[htb]
\label{table:ddim}
    \vspace{-0.5cm}
    \caption{Quantitative evaluation with different DDIM step in Linear Inverse Problem}
    \centering
    \footnotesize

    \tabcolsep=0.22cm
    \resizebox{\textwidth}{!}{
    \begin{tabular}{c |  ccc | ccc | ccc | c | }
        \toprule[1.2pt]
          \multirow{2}{*}{Methods}  &  \multicolumn{3}{c}{Inpainting} \vline & \multicolumn{3}{c}{Super resolution} \vline & \multicolumn{3}{c}{Gaussian deblurring} \vline & \multirow{2}{*}{DDIM steps}   \\
        \rule{0pt}{8pt}
            & SSIM$\uparrow$ & PSNR$\uparrow$ & LPIPS↓ & SSIM$\uparrow$ & PSNR$\uparrow$ & LPIPS↓ & SSIM$\uparrow$ & PSNR$\uparrow$ & LPIPS↓ &\\
        \toprule[1.2pt]
        
        DPS~\cite{dps} & 0.815 &  27.73 & 0.233 & 0.649 & 22.99 & 0.329 & 0.662 & 23.43 & 0.310 &  100\\

        DPS+DSG(Ours) & \textbf{0.899} & \textbf{30.95} & \textbf{0.132} & \textbf{0.784} & \textbf{26.82} & \textbf{0.227} & \textbf{0.762} & \textbf{26.16} & \textbf{0.239} & 100\\

        \toprule[1.2pt]
        DPS~\cite{dps} &  0.704	& 22.74 & 0.314 & 0.568 & 19.57 & 0.405 & 0.543 & 18.59 & 0.428 &  50 \\
        
        DPS+DSG(Ours)  &  \textbf{0.886} & \textbf{30.67} & \textbf{0.176} & \textbf{0.790} & \textbf{26.96} & \textbf{0.231} & \textbf{0.749} & \textbf{25.76} & \textbf{0.264} & 50 \\
        \toprule[1.2pt]

        DPS~\cite{dps} &  0.408 & 12.08 & 0.590 & 0.337 & 10.93 & 0.614 & 0.353 & 10.97 & 0.639 &  20 \\

        DPS+DSG(Ours)  & \textbf{0.812} & \textbf{27.53} & \textbf{0.267} & \textbf{0.771} & \textbf{26.68} & \textbf{0.272} &\textbf{ 0.771} & \textbf{26.68} & \textbf{0.272} & 20 \\
        \bottomrule[1.2pt]
    \end{tabular}
    }
    \vspace{-0.5cm}
    \label{tb:ddim}
\end{table*}

\textbf{Style Guidance} For Style Guidance, our test set consists of 1,000 paintings by various artists collected from WikiArt \cite{wikiart}, and we employ LDM \cite{stable} as the pre-trained diffusion prior. Freedom \cite{freedom} and LGD \cite{Lossguided} are chosen as the baseline. For an input image $x_{in}$, the loss guidance can be represented as:
\begin{equation}
    \text{Loss}(\hat{x}_0(x_t), x_{in}) = ||E(\hat{x}_0(x_t)) - E(x_{in})||_F^2,
\end{equation}
where $E(\cdot)$ is the Gram matrix of the 3rd feature map extracted from the CLIP \cite{clip} image encoder and $||\cdot||_F^2$ denotes the Frobenius norm. 

\begin{figure}[htbp]
  \centering
  \vspace{-0.2cm}
  \includegraphics[width=0.95\linewidth]{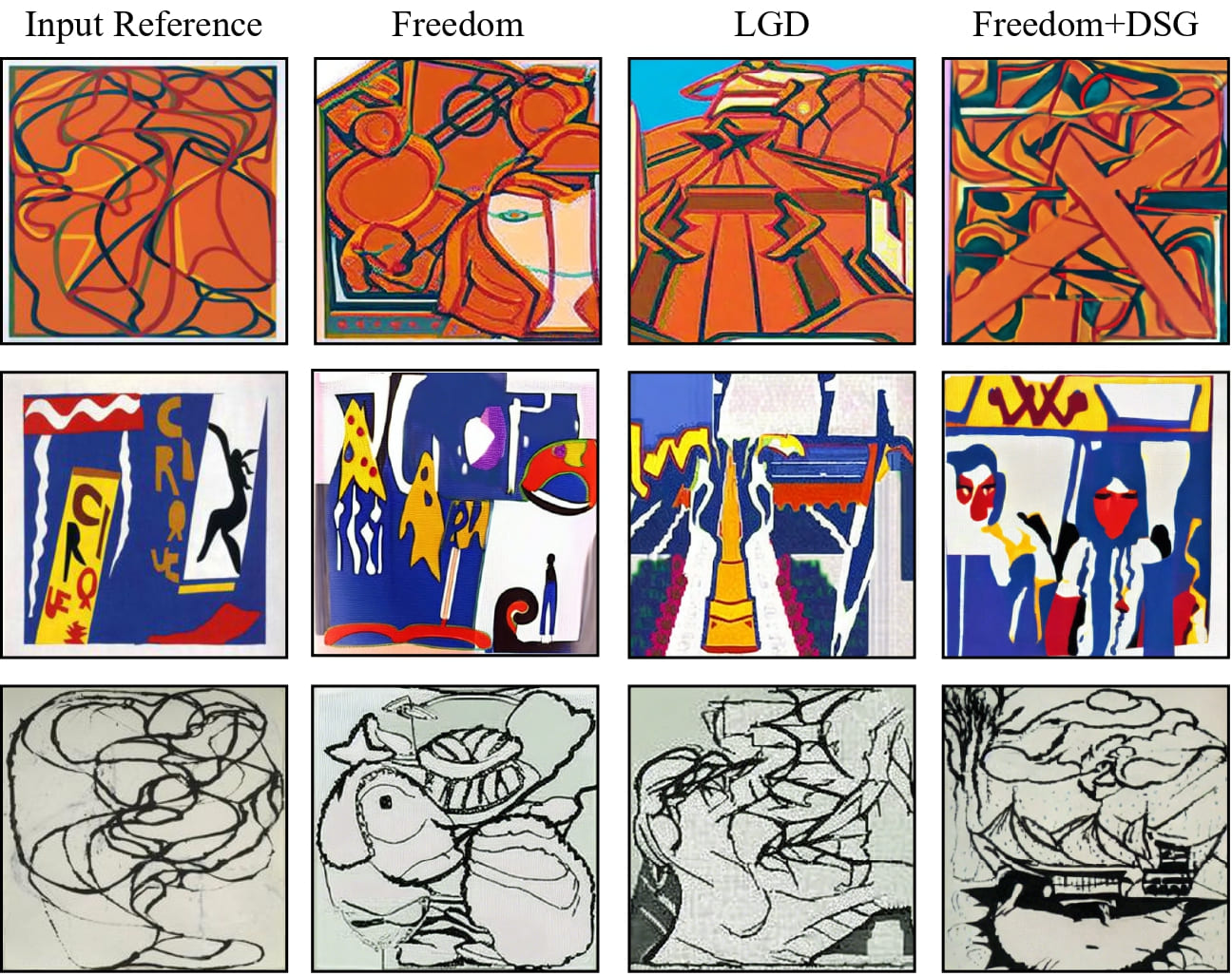}
  \caption{Qualitative results of Style Guidance using pre-trained Stable Diffusion.}
  \label{fig:style}
\end{figure}

\textbf{Text-Segmentation Guidance} In the context of Text-Segmentation Guidance, we apply our method to a more challenging task, incorporating multiple conditions to guide the conditional generation process. For a given text description $t$ and a segmentation map $M$ as input, the objective is to generate an image $x_0$ that aligns with both the input text $t$ and the provided segmentation mask $M$ within 500 denoising steps. To achieve this, we utilize LDM \cite{stable} as the diffusion prior, allowing us to integrate the original text conditions through its conditional denoiser, and use the additional plug-and-play segmentation guidance:
\begin{equation}
    \text{Loss}(\hat{x}_0(x_t)), M) = CE(E(\hat{x}_0(x_t)),M).
\end{equation}
Here $CE(\cdot)$ represents the Cross-Entropy Loss and $E(\cdot)$ represents the MobileNetV3-Large segmentation network. 

\begin{figure}[htbp]  
\includegraphics[width=1\linewidth]{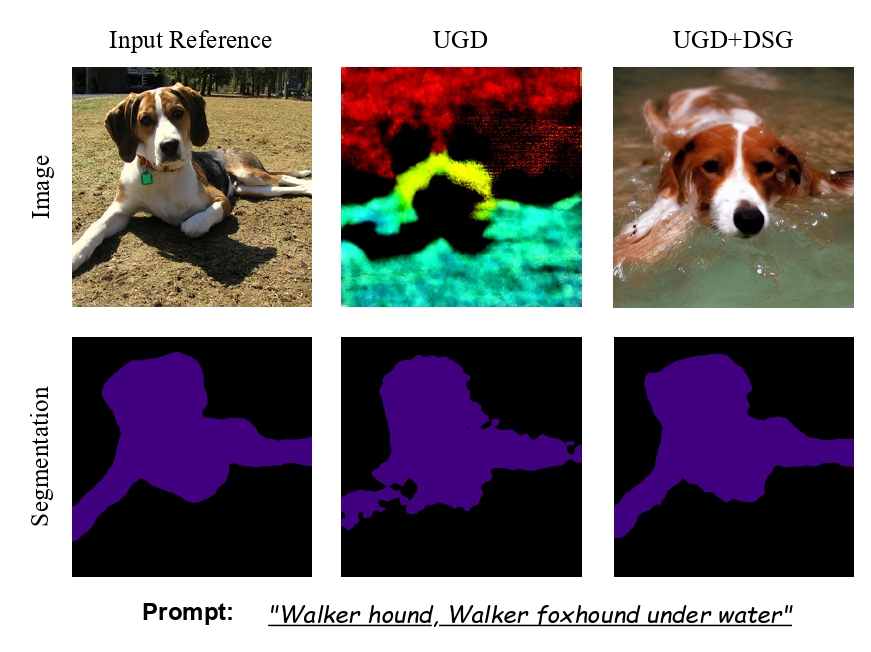}
  \vspace{-0.9cm}
  \caption{Qualitative result of Text-Segmentation Guidance using 500 denoising steps with Stable Diffusion. }
  \vspace{-0.5cm}
\label{fig:text} 
\end{figure}

\textbf{FaceID Guidance} In FaceID Guidance, we utilized a pre-trained diffusion model from CelebA-HQ256*256 provided by \cite{freedom} and employed the CelebA-HQ test set as the input reference image for generating 1k images in 100 denoising steps. For an input reference image $I$, we use the FaceID loss for guidance:
\begin{equation}
    \text{Loss}(\hat{x}_0(x_t)), I) = CE(E(\hat{x}_0(x_t)),E(I)),
\end{equation}
where $E(\cdot)$ represents ArcFace \cite{arcface}, a human face recognition network used to extract features from the facial image.

\textbf{Text-Style Guidance} In Text-Style Guidance, for a given text description $t$ and a reference image $I$ as input, we utilize LDM \cite{stable} as the pre-trained conditional diffusion prior and utilize the style loss for guidance, which is the same as Style Guidance:
\begin{equation}
    \text{Loss}(\hat{x}_0(x_t), x_{in}) = ||E(\hat{x}_0(x_t)) - E(x_{in})||_F^2,
\end{equation}
Please refer to the Appendix \ref{sec:addexp} for further experimental details and additional results for FaceID Guidance and Text-Style Guidance. 

\vspace{-0.15cm}
\subsection{Comparison with the State-of-the-art}

\begin{figure}[htbp]
  \centering
  \vspace{-0.2cm}
  \includegraphics[width=0.9\linewidth]{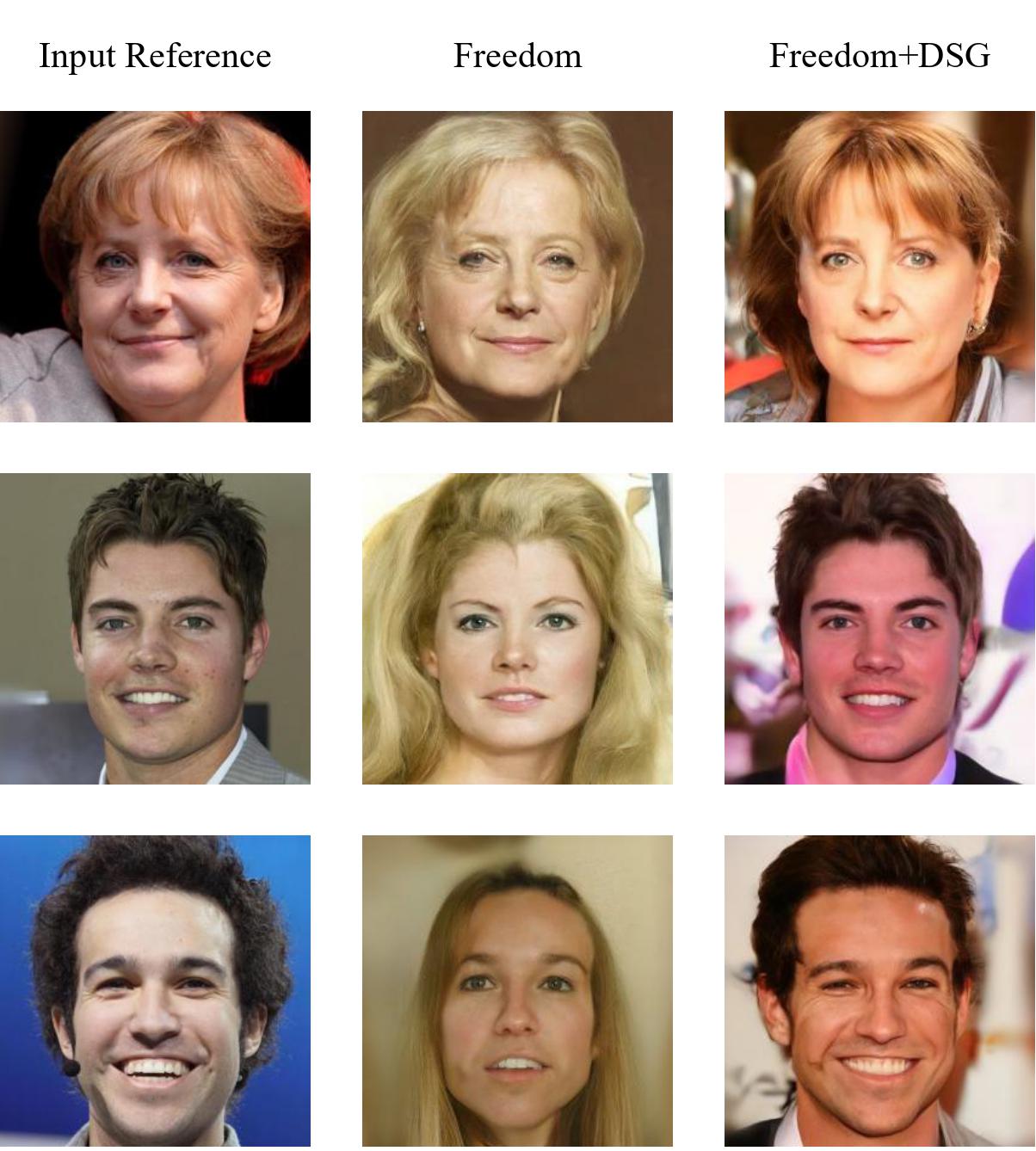}
   \vspace{-0.4cm}
  \caption{The qualitative results in FaceID Guidance using a diffusion model pre-trained from CelebA-HQ256*256.}
  \vspace{-0.2cm}
\label{fig:faceid1} 
\end{figure}

The qualitative and quantitative results are shown in Table \ref{table:comparison}, \ref{table:faceid} and Figure \ref{fig:lin}, \ref{fig:style}, \ref{fig:text}, \ref{fig:faceid1}. The result in Table \ref{table:comparison}, \ref{table:faceid} demonstrates that our method outperforms the current state-of-the-art training-free method by a significant margin. Our method produces samples that exhibit both increased realism and better alignment with the input conditions. 

In the context of linear inverse problems, our method excels in preserving intricate details, such as facial features and hair, whereas DPS tends to apply smoothing and blurring effects to these details. With FaceID Guidance, our model generates faces that more closely match the input reference compared to baseline methods. For Style Guidance, Text-Segmentation guidance, and Text-Style Guidance, the images generated by our method also exhibit improved alignment with the input conditions while reducing artifacts. 
\begin{table}[h]
    \caption{The quantitative result in CelebA-HQ test set for FaceID guidance}

    \label{table:faceid}
    \centering
    \small
    \vspace{0.2cm}
    \tabcolsep=0.22cm

    \begin{tabular}{|c |  c | c | c|}
        \toprule[1.2pt]
         Method  & FaceID & FID & KID \\
        \toprule[1.2pt]
        Freedom  & 0.545 & 39.61 & 0.0165 \\
        LGD  & 0.533 & 39.13 & 0.0170 \\
        \toprule[1.2pt]
        Freedom+DSG  & \textbf{0.371} & \textbf{34.29} & \textbf{0.0092} \\  
        \bottomrule[0.8pt]
    \end{tabular}
    
    \label{tb:faceid}
\end{table}

\subsection{Ablation Study}
\label{sec:ablation}

\begin{figure}[htbp]
  \centering
  \vspace{-0.1cm}
  \includegraphics[width=1\linewidth]{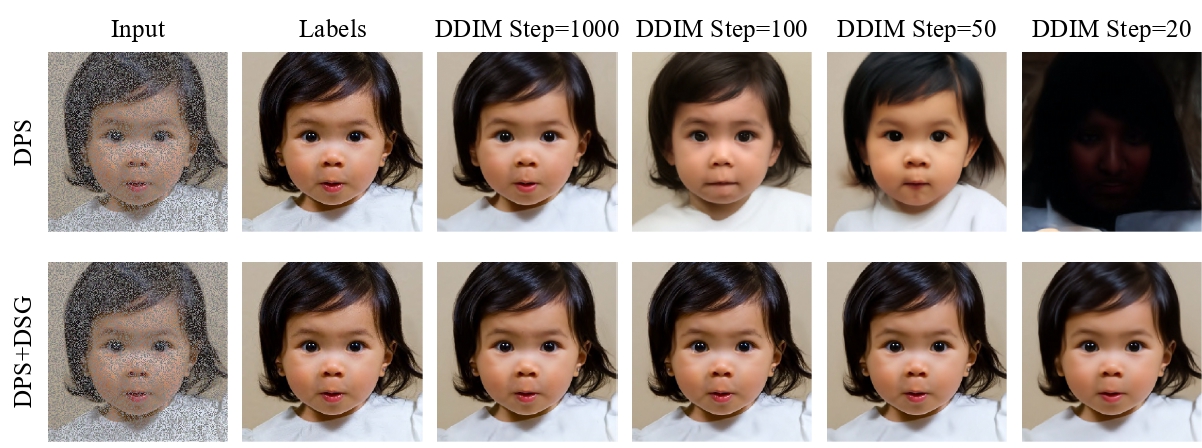}
  \caption{Comparison between DPS and DPS+DSG in different denoising steps in solving inpainting. With the decreasing of denoising steps, DPS can not align with the input due to the small step size. However, when plugging in DSG, it can allow larger step sizes, thus achieving better alignment.}
  \vspace{-0.3cm} 
  \label{fig:ddim}
\end{figure}

\textbf{Robustness with fewer denoising steps} When reducing the number of denoising steps (Table \ref{tb:ddim} and Figure \ref{fig:ddim}), we observe a more pronounced performance gap.  This phenomenon can be attributed to the limitations of DPS, which restricts the small step sizes for not deviating far from the intermediate manifold $\mathcal{M}_t$. Consequently, with fewer guidance steps, aligning with the measurements becomes increasingly challenging. In contrast, our model experiences only a slight decrease in performance. This can be attributed to our approach, which allows for larger step sizes while still preserving the underlying manifold structure. As a result, even with a reduced number of denoising steps, we can still achieve accurate alignment with the measurements while generating realistic samples, as shown in Figure \ref{fig:ddim}. Please refer to Appendix \ref{app:ddim_setting} for further details.

\begin{figure}[htbp]
  \centering
  \vspace{-0.2cm}
  \includegraphics[width=0.95\linewidth]{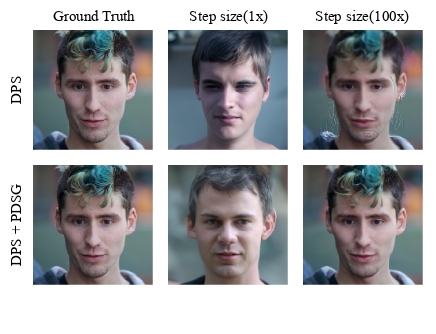}
   \vspace{-1cm}
  \caption{Comparison between DPS and DPS+PDSG (Another version of DSG by projecting DPS into Spherical Gaussian constraint) using 50 denoising steps.}
  \vspace{-0.2cm}
\label{fig:pdsg} 
\end{figure}

\textbf{Advantanges of guidance with Spherical Gaussian constraint.} We performed additional comparative experiments to demonstrate the advantages of constraining the guidance with the spherical Gaussian constraint. To demonstrate the benefits of spherical Gaussian constraint, we implemented another version of DSG(PDSG) by projecting the $x_{t-1}$ obtained by DPS onto the hyperphere $S^n_{\mu_\theta(x_t),\sqrt{n}\sigma}$ in Equation \ref{eq:dsg2}. With this projection, DPS can have a larger step size(100x) than the original setting with no artifacts. See Appendix \ref{app:dps_proj} for further details.

\section{Limitations}
Although DSG addresses the manifold deviation problem, allowing for larger guidance steps that enhance time efficiency, its sampling strategy may compromise sample diversity. This arises for two main reasons: Firstly, DSG restricts Gaussian sampling to the proximity of the gradient descent direction, while baseline methods (such as DPS) sample random noise across all directions. Secondly, approximating the high-dimensional Gaussian in DSG to a hypersphere could potentially diminish sample diversity, even though the possibility of sampling points outside the hypersphere is very small.

\section{Conclusion}
In this paper, we have revealed a crucial issue in the training-free conditional diffusion models: the occurrence of manifold deviation during the sampling process when employing loss guidance. This phenomenon is substantiated by establishing a certain lower bound for the estimation error of the loss guidance. To tackle this issue, we have proposed Diffusion with Spherical Gaussian constraint (DSG), inspired by the concentration phenomenon in high-dimensional Gaussian distributions. DSG effectively constrains the guidance step within the intermediate data manifold through optimization, thereby mitigating the manifold deviation problem and enabling the utilization of larger guidance steps. Furthermore, we presented a closed-form solution for DSG denoising with the Spherical Gaussian constraint. It's worth noting DSG serves as a plug-and-play module for training-free
conditional diffusion models (CDMs). Integrating DSG into these CDMs merely involves modifying a few lines of code with almost no extra computational cost, but yields significant performance improvements. We have integrated DSG into several recent new CDMs for various conditional generative tasks. The experimental results validate the superiority and adaptability of DSG in terms of both sample quality and time efficiency. 

Notably, the improvements of DSG on both sample quality and time efficiency come at the expense of sample diversity. This is because DSG replaces the random Gaussian noise term in the reverse diffusion process with the deterministic conditional guidance step. Although we use the guidance rate $g_r$ to alleviate this problem in practice, it cannot be avoided inherently. Enhancing DSG diversity while preserving its generation quality is under consideration in our future work. 

\section*{Acknowledgement}

This work was supported by NSFC (No.62303319), Shanghai Sailing Program (22YF1428800, 21YF1429400), Shanghai Local College Capacity Building Program (23010503100), Shanghai Frontiers Science Center of Human-centered Artificial Intelligence (ShangHAI), MoE Key Laboratory of Intelligent Perception and Human-Machine Collaboration (ShanghaiTech University), and Shanghai Engineering Research Center of Intelligent Vision and Imaging. Additionally, we would like to express our sincere gratitude to Professor Xuming He for his invaluable suggestions during the revision of this paper.

\clearpage
\section*{Impact Statement}
There are many potential societal consequences of our work, none of which we feel must be specifically highlighted here. 
\bibliography{reference}
\bibliographystyle{icml2024}

\clearpage
\appendix
\onecolumn

\section{Proofs}

\subsection{Proof of Proposition 4.1}

\begin{proposition}
(\textit{Lower bound of Jensen gap}). For the $\beta$-strongly convex function $f$ and the random variable $X\in \mathbb{R}^n \sim \mathcal{N}(\mu, \Sigma)$, we can have the lower bound of the Jensen Gap:
\[
    \mathcal{J} \ge \frac{1}{2}\beta \sum_{i=1}^n \lambda_i.
\]
\end{proposition}

\textbf{Proof:}
Since $f(x)$ is a $\beta$-strongly convex function, we have
\[
    \begin{gathered}
        f(\frac{1}{2} (x_1 + x_2)) \le \frac{1}{2} \left[f(x_1) +  f(x_2)\right] - \frac{\beta}{8} \|x_1-x_2\|^2, \\ 
    \end{gathered}
\]
Then, for convenience, let random variable $\delta=x-\mu$, and $\delta \sim \mathcal{N}(0, \Sigma)$. we obtain
\[
    \begin{aligned}
        \mathcal{J}(f,x \sim \mathcal{N}(\mu, \Sigma)) = & \mathbb{E}\left[f(x)\right] - f(\mathbb{E}\left[x\right]) \\
        = & \int p(x)f(x)dx-f(\mu) \\
        =& \frac{1}{2}\int p(\mu+\delta) f(\mu+\delta) d\delta + \frac{1}{2}\int p(\mu-\delta) f(\mu-\delta) d\delta - f(\mu) \\
        = & \int p(\mu + \delta) \left(\frac{1}{2}f(\mu+\delta)+\frac{1}{2}f(\mu-\delta)-f(\mu)\right) d\delta \quad 
        \diamondsuit \text{ symmetry of Gaussian distribution} \\
        \ge & \frac{\beta}{2} \int p(\mu + \delta)  \|\delta\|^2 d\delta \quad \diamondsuit \text{ $\beta$-strongly convexity}\\
        =& \frac{\beta}{2}\mathbb{E}\left[\|\delta\|^2\right].
    \end{aligned}
\]
Since here the covariance matrix $\Sigma$ is symmetric, we can use the spectral theorem here and write $\Sigma = P^T\Lambda P$ where $P$ is an orthogonal matrix and $\Lambda$ is diagonal with positive diagonal elements $\lambda_1,\cdots,\lambda_n$. Then, let $z=P\Sigma^{-\frac{1}{2}}\delta$ and $z\sim \mathcal{N}(0, I)$. Thus, we have 
\[
    \|\delta\|^2=\delta^T\delta = \left(\Sigma^{-\frac{1}{2}}\delta\right)^T \Sigma \left(\Sigma^{-\frac{1}{2}}\delta\right)=\left(\Sigma^{-\frac{1}{2}}\delta\right)^T P^T\Lambda P \left(\Sigma^{-\frac{1}{2}}\delta\right)=\|z\|^2_\Lambda
\]
Consider each component in $z$ is independent.
\[
    E\left[\|\delta\|^2\right] = \mathbb{E}\left[\|z\|^2_\Lambda\right] = \sum_{i=1}^n \mathbb{E}\left[ \lambda_i z_i^2\right] = \sum_{i=1}^n \lambda_i.
\]

Therefore, we have the lower bound of the Jensen gap that
\[
    \mathcal{J} \ge \frac{1}{2}\beta \sum_{i=1}^n \lambda_i.
\]

\subsection{Proof of Proposition 4.3}
\label{app:gaussian}

\begin{proposition}
(Concentration of high-dimensional isotropy Gaussian distribution). For any n-dimensional isotropy Gaussian distribution $x \sim N(\mu,\sigma^2 I)$:
\begin{equation}
    \begin{split}
        P(||x-\mu||_2^2 \geq  x_{lower}=n\sigma^2+2n\sigma^2 (\sqrt{\epsilon}+\epsilon)) \leq e^{-n\epsilon}, \nonumber \\ 
        P(||x-\mu||_2^2 \leq x_{upper}=n\sigma^2 - 2n\sigma^2\sqrt{\epsilon}) \leq e^{-n\epsilon}. \nonumber\\
    \end{split}
\end{equation}
When $n$ is sufficiently large, it is close to the uniform distribution on the hypersphere of radius $\sqrt{n}\sigma$.
\end{proposition}

\textbf{Proof:}

First of all, we can find the expected value of the $||x-\mu||_2^2$ in the following way:

\begin{equation}
    E(||x-\mu||_2^2) =  \sum_{i=1}^{n} E((x_i-\mu)^2) = \sum_{i=1}^{n} E^2((x_i-\mu)) + D(x_i-\mu) = n\sigma^2,
\end{equation}

where $x_i$ is $ith$ element of $x$.

Then, the concentration of $||x-\mu||_2^2$ can be proved using \textbf{standard Laurent-Massart bound} for a chi-square
distribution. If $Z$ is a chi-square distribution with $n$ degrees of freedom,

\begin{equation}
    \begin{split}
        P[Z-n \geq 2\sqrt{nt}+ 2t] \leq e^{-t}, \\
        P[Z-n \leq -2\sqrt{nt}] \leq e^{-t}. \\
    \end{split}
\end{equation}

By substituting $Z=\sum_{i=1}^{n} (x_i-\mu)^2 / \sigma^2$ and $t=n\epsilon$,

\begin{equation}
    \begin{split}
        P(\frac{1}{\sigma^2} \sum_{i=1}^{n} (x_i-\mu)^2  \geq n + 2n(\sqrt{\epsilon}+\epsilon)) \leq e^{-n\epsilon}, \\
        P(\frac{1}{\sigma^2} \sum_{i=1}^{n} (x_i-\mu)^2  \leq n -2n\sqrt{\epsilon}) \leq e^{-n\epsilon}. \\
    \end{split}
\end{equation}

Therefore,

\begin{equation}
    P(r\sqrt{1-2\sqrt{\epsilon}} \leq ||x-\mu||_2 \leq r\sqrt{1+2\sqrt{\epsilon}+2\epsilon}) \geq 1- 2e^{-n\epsilon}.
\end{equation}

By choosing $\epsilon_{max} = max\{1-\sqrt{1-2\sqrt{\epsilon}},\sqrt{1+2\sqrt{\epsilon}+2\epsilon}-1\}$,

\begin{equation}
    P((1-\epsilon_{max})r \leq ||x-\mu||_2 \leq (1+\epsilon_{max})r) \geq 1- 2e^{-n\epsilon},
\end{equation}

which proves the concentration of $||x-\mu||_2$.

\subsection{Closed-form Solution of Equation 13}
\label{app:solution}

Given the optimization problem
\begin{equation}
    \begin{split}
        \mathop{\arg\min}\limits_{x'} & \left[\nabla_{x_t} L(\hat{x}_0(x_t),y)\right]^T (x'-x_t)  \\
        \text{s.t.}& \;x' \in S^d_{\mu_\theta(x_t),\sqrt{n}\sigma_t},
    \end{split}
\end{equation}
where $S^d_{\mu,r}=S^d_{\mu_\theta(x_t),\sqrt{n}\sigma}=\{x: ||x-\mu_\theta(x_t)||_2^2=n \sigma_t^2 \}$, the optimal solution can be derived as follows:
\begin{equation}
    \begin{split}
        &\mathop{\arg\min}\limits_{x'}  \left[\nabla_{x_t} L(\hat{x}_0(x_t),y)\right]^T (x'-x_t) \\
        =& \mathop{\arg\min}\limits_{x'} \left[\nabla_{x_t} L(\hat{x}_0(x_t),y)\right]^T ((x'- \mu_\theta(x_t))+(\mu_\theta(x_t) -x_t)) \\
        =& \mathop{\arg\min}\limits_{x'} \left[\nabla_{x_t} L(\hat{x}_0(x_t),y)\right]^T (x'- \mu_\theta(x_t)). \\
    \end{split}
\end{equation}
By reparameterizing $x'=\mu_\theta(x_t)+\sqrt{n}\sigma_t d$ where $||d||_2^2=1$ using constraint,
\begin{equation}
    \begin{split}
        &\mathop{\arg\min}\limits_{x'}  \left[\nabla_{x_t} L(\hat{x}_0(x_t),y)\right]^T (x'-x_t) \\
        =& \mathop{\arg\min}\limits_{d} \sqrt{n}\sigma_t \left[\nabla_{x_t} L(\hat{x}_0(x_t),y)\right]^T d.
    \end{split}
\end{equation}
Obviously, when $d = -\nabla_{x_t} L(\hat{x}_0(x_t),y) / ||\nabla_{x_t} L(\hat{x}_0(x_t),y)||_2$, the optimization problem gets the minimal value $-\sqrt{n}\sigma_t ||\nabla_{x_t} L(\hat{x}_0(x_t),y)||_2$.

Therefore, $x^*_{t-1}=x'= \mu_\theta(x_t) - \sqrt{n}\sigma_t \nabla_{x_t} L(\hat{x}_0(x_t),y) / ||\nabla_{x_t} L(\hat{x}_0(x_t),y)||_2$.

\begin{table*}[t]
    \caption{Hyperparameter settings of DSG for Linear Inverse Problems in FFHQ}

    \centering
    \normalsize
    \tabcolsep=0.22cm
    \begin{tabular}{c |  c | c | c}
        \toprule[1.2pt]
         Task  & Interval & Guidance Rate & Denoising steps \\

        \toprule[1.2pt]

        Inpainting  & 5 & 0.2 & 1000 \\
        Super-Resolution  & 20 & 0.2 & 1000 \\
        Gaussian-deblurring  & 5 & 0.2 & 1000 \\
        \hline
        Inpainting  & 1 & 0.2 & 100 \\
        Super-Resolution  & 2 & 0.1 & 100 \\
        Gaussian-deblurring  & 1 & 0.1 & 100 \\
        \hline
        Inpainting  & 1 & 0.2 & 50 \\
        Super-Resolution  & 1 & 0.1 & 50 \\
        Gaussian-deblurring  & 1 & 0.1 & 50 \\
        \hline
        Inpainting  & 1 & 0.2 & 20 \\
        Super-Resolution  & 1 & 0.2 & 20 \\
        Gaussian-deblurring  & 1 & 0.2 & 20 \\
        
        \bottomrule[0.8pt]
    \end{tabular}
    
    \label{tb:hyper}
\end{table*}

\begin{figure*}[htbp]
  
  \centering
  \includegraphics[width=0.9\linewidth]{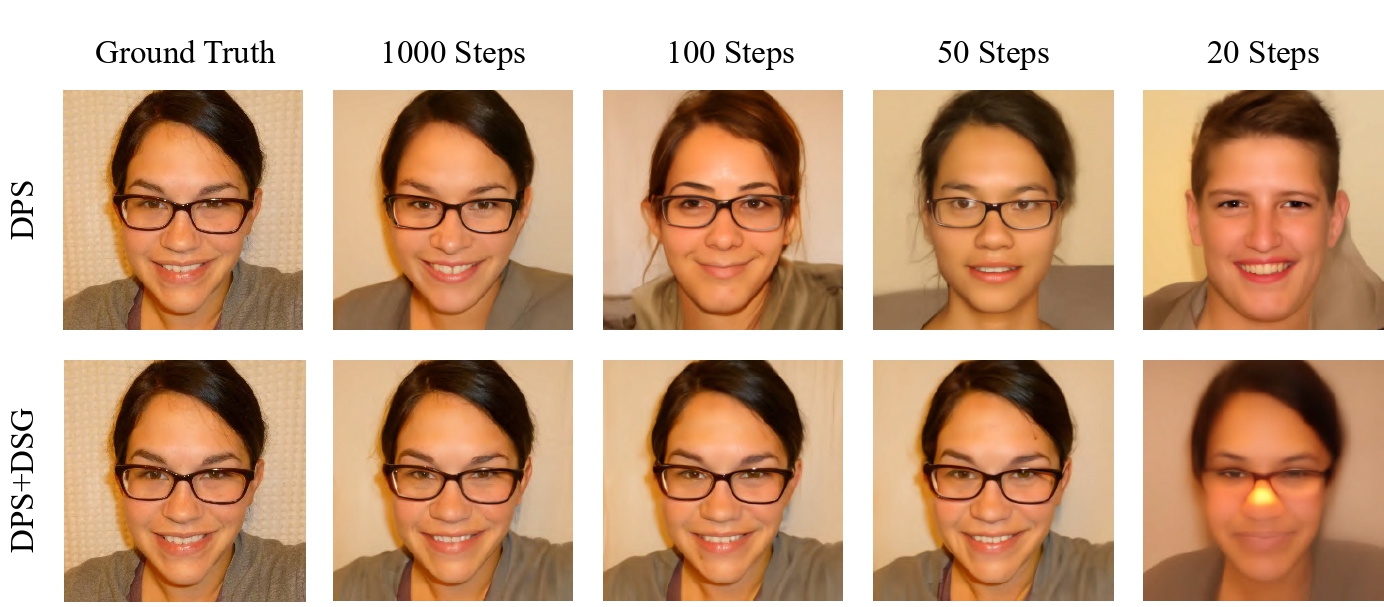}
  \caption{Comparison between DPS and DPS+DSG in Super-resolution task using different denoising steps.}
  \label{fig:ddim_app}
\end{figure*}

\begin{figure*}[t]

  \centering
  \includegraphics[width=0.9\linewidth]{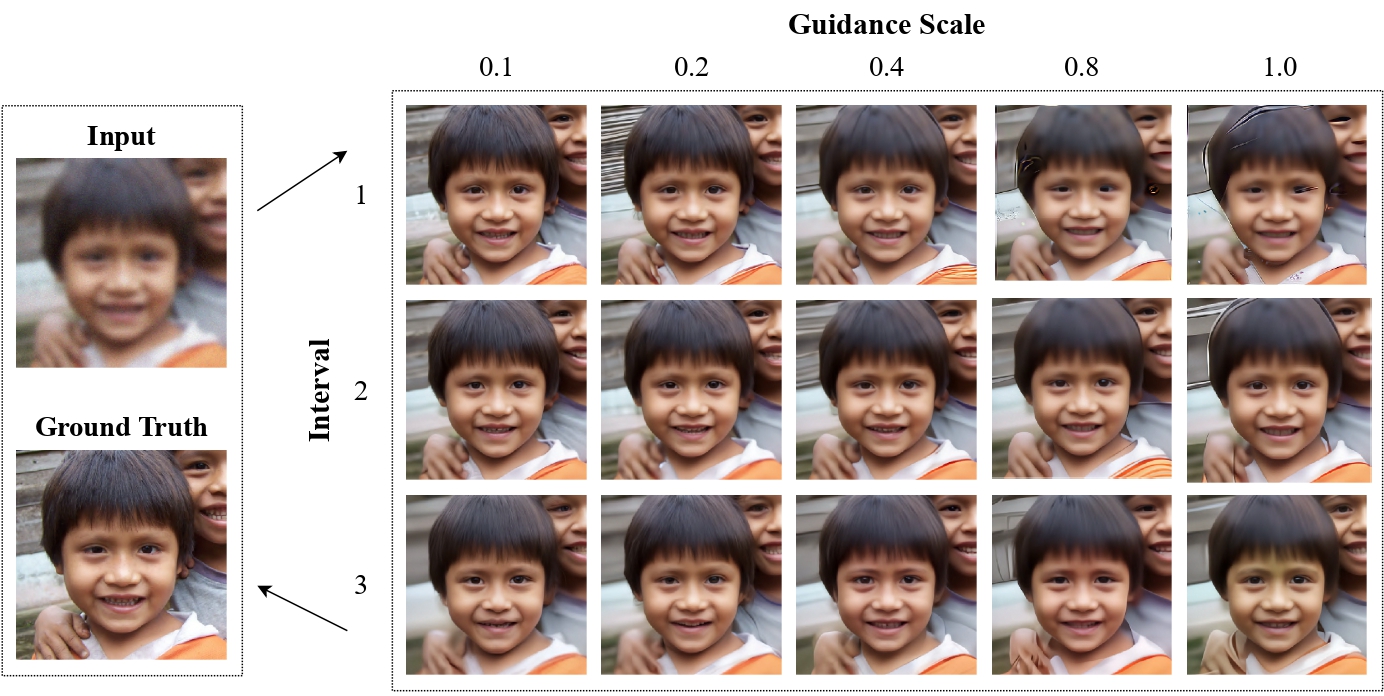}
  \caption{Different guidance rate using DSG using 100 DDIM steps}
  \label{fig:hyper}
\label{fig5} 
\end{figure*}

\section{Ablation Study}
\subsection{Hyperparameter Analysis}
Our model has two main hyperparameters: guidance rate $g_r$ and interval $i$. The guidance rate represents the weight of guidance in Equation \ref{eq:mix}. When it is zero, the denoising process is equivalent to unconditional generation. When it is set to 1, then the path of generation is determined. In practice, we find it better to choose values between 0.05 and 0.2 because a certain level of random noise is a trade-off between unconditional sampling diversity and better alignment. The interval means that we apply guidance at a fixed interval and do unconditional sampling when not applying guidance. It can also increase the diversity of DSG and usually be used when the denoising steps are large, e.g. denoising steps larger than 100.

The hyperparameters we used for the linear inverse problem in FFHQ are shown in Table \ref{tb:hyper}. For the linear inverse problem in ImageNet, we use $g_r=0.2,i=5$ for Inpainting, $g_r=0.1,i=10$ for Super-resolution, and $g_r=0.1,i=5$ for Gaussian deblurring. For Style Guidance, Text-Style Guidance and Text-Segmentation Guidance, we set $g_r=0.1,i=1$. For FaceID Guidance, we set the $g_r=0.05,i=1$.

From Table \ref{tb:hyper}, we can observe that when the number of denoising steps is limited($\leq100$), simply setting the interval to 1 and the guidance rate to a small value ([0.05, 0.2]) can yield a satisfactory result. When denoising steps are significantly large (e.g., $T=1000$), a large interval (larger than 5) can be set to increase the diversity of our method and decrease the number of guidance, thus accelerating the inference time while enhancing the quality of the generated image. 

It is also worth noticing that while other training-free methods require careful tuning of the step size with different loss functions, the step sizes of DSG are adaptive and independent of the loss function. Moreover, the similar hyperparameter settings of DSG \textbf{can be applied to almost all tasks}, which reduces the cost for hyperparameter searching.

\subsection{Ablation Study with Different Denoising Steps Compared with DPS}
\label{app:ddim_setting}

When conducting the ablation study of different denoising steps in the linear inverse problem, we carefully tune the step size $\gamma$ of DPS \cite{dps}  to 10$\times$ than the original setting with denoising steps=100 and 20$\times$ with denoising steps=20, 50. Further results are shown in Figure \ref{fig:ddim_app}.

\subsection{Ablation Study on Projecting DPS into Spherical Gaussian}
\label{app:dps_proj}

To further demonstrate the advantages of Spherical Gaussian constraint,  we first consider the single denoising step from $x_t$ to $x_{t-1}$, DPS calculates $x_{t-1}$ by estimating the additional correction step:
\begin{equation}
    x_{t-1} = \underbrace{DDIM(x_t, \epsilon_\theta(x_t,t))}_{\text{sampling step}} - \underbrace{\gamma \nabla_{x_t} L(\hat{x}_0(x_t),y)}_{\text{correction step}}.
\end{equation}
However, when $\gamma$ is pretty large. we project $x_{t-1}$ obtained by DPS onto the hyperplane $S_{\mu_\theta (x_t),\sqrt{n}\sigma}^n$ to force the spherical Gaussian constraint, which is called DPS+PDSG. The projection process can be represented as:
\begin{equation}
    d_p =  x_{t-1} - \mu_\theta (x_t),
    \label{appendix:proj1}
\end{equation}
\begin{equation}
x^p_{t-1} = \mu_\theta (x_t) + \sqrt{n}\sigma_t d_p / ||d_p||,
 \label{appendix:proj2}
\end{equation}
where $x^p_{t-1}$ denates the projection point, which is the closest point in $S_{\mu_\theta (x_t),\sqrt{n}\sigma}^n$ w.r.t. $x_{t-1}$. Since a large step size $\gamma$ will cause DPS to fall off the manifold (Figure \ref{fig:ddim_app}), the operation in Equation \ref{appendix:proj1}, \ref{appendix:proj2} can project the $x_{t-1}$ obtained by DPS back into the intermediate manifold $\mathcal{M}_{t-1}$ while allowing for larger step sizes.

\subsection{Ablation Study on Denoising Process}
\begin{figure*}[htbp]
  
  \centering
  \includegraphics[width=0.9\linewidth]{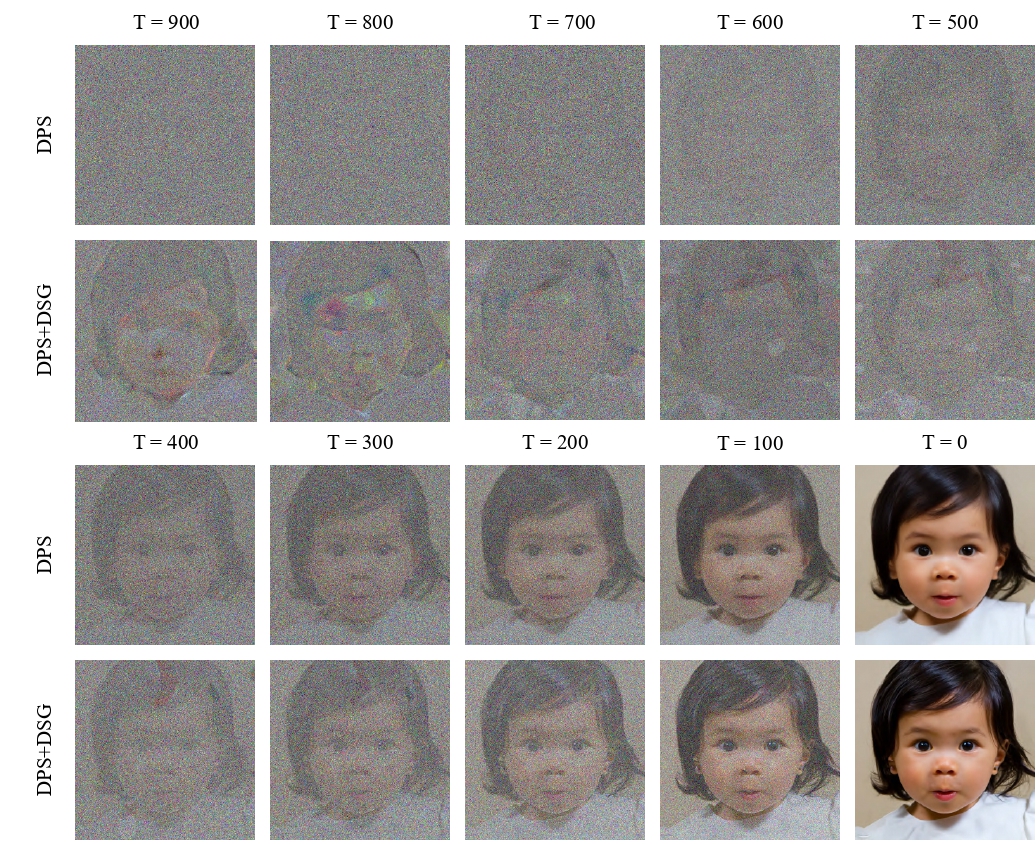}
  \caption{Qualitative results of the denoising process. }
  \label{appendix:denoise}
\end{figure*}

We compare our DSG with DPS in the Inpainting task in 1000 denoising steps and show the intermediate noisy image $x_t$ in Figure \ref{appendix:denoise}. The results demonstrate that our guidance is more effective and can expedite the restoration of the overall appearance of the image.

\section{Additional Experimental Details and Results}
\label{sec:addexp}

In this section, to further demonstrate the applicability and strength of our DSG, we will provide additional experimental details and results.

\subsection{More Experimental Details and Results for FaceID Guidance}
\label{sec:app_faceid}

\begin{figure*}[htbp]
  \centering
  
  \includegraphics[width=0.9\linewidth]{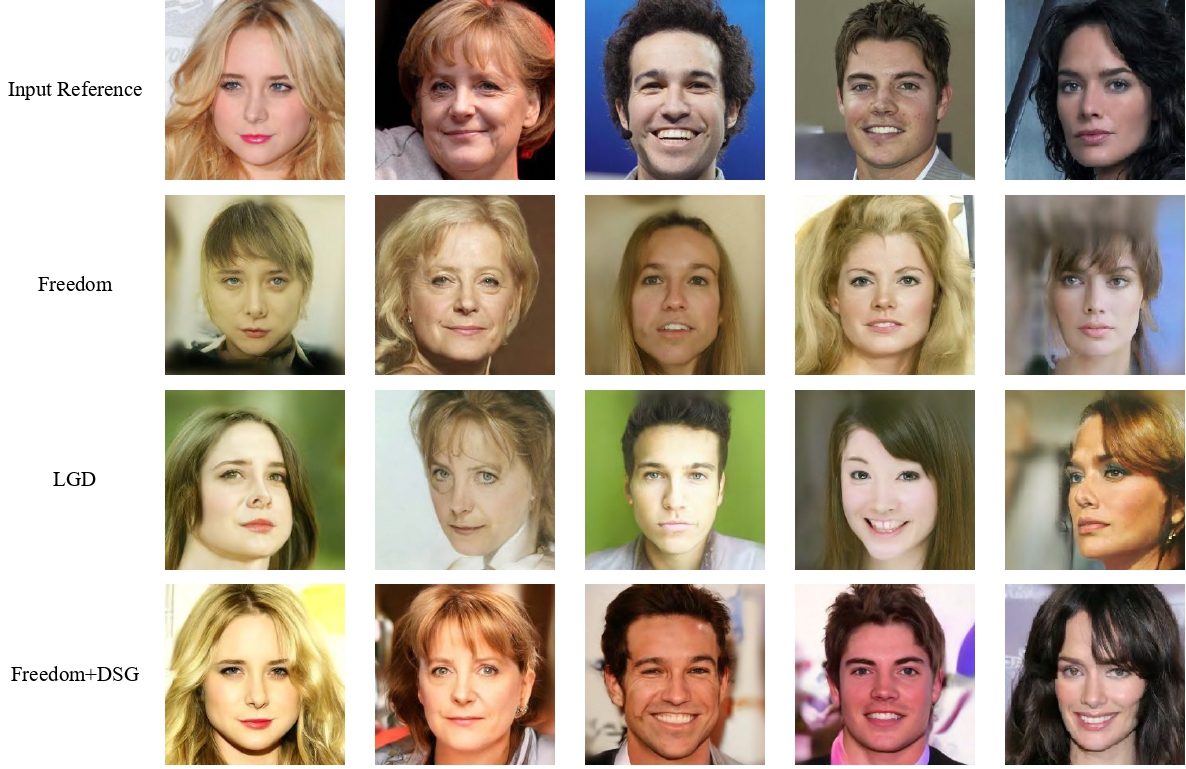}
  \caption{The qualitative results in FaceID Guidance using a diffusion model pre-trained from CelebA-HQ256*256.}
  \label{fig:faceid}
\end{figure*}

In FaceID Guidance, we choose Freedom \cite{freedom}, LGD \cite{Lossguided} as the baselines. For Freedom, we follow the hyperparameter settings proposed in the original paper. For LGD, we set the number of Monte Carlo samples to 3, covariance to $0.5\sqrt{1-\alpha_t}$, and step size to $100\sqrt{\alpha_t}$. As shown in Table \ref{table:faceid}, Figure \ref{fig:faceid1} and Figure \ref{fig:faceid}, our DSG shows the SOTA performance according to these qualitative and quantitative results.

\subsection{More Experimental Details and Results for Text-Style Guidance}
\label{sec:app_text_style}

\begin{figure*}[htbp]
  \centering
  \includegraphics[width=0.9\linewidth]{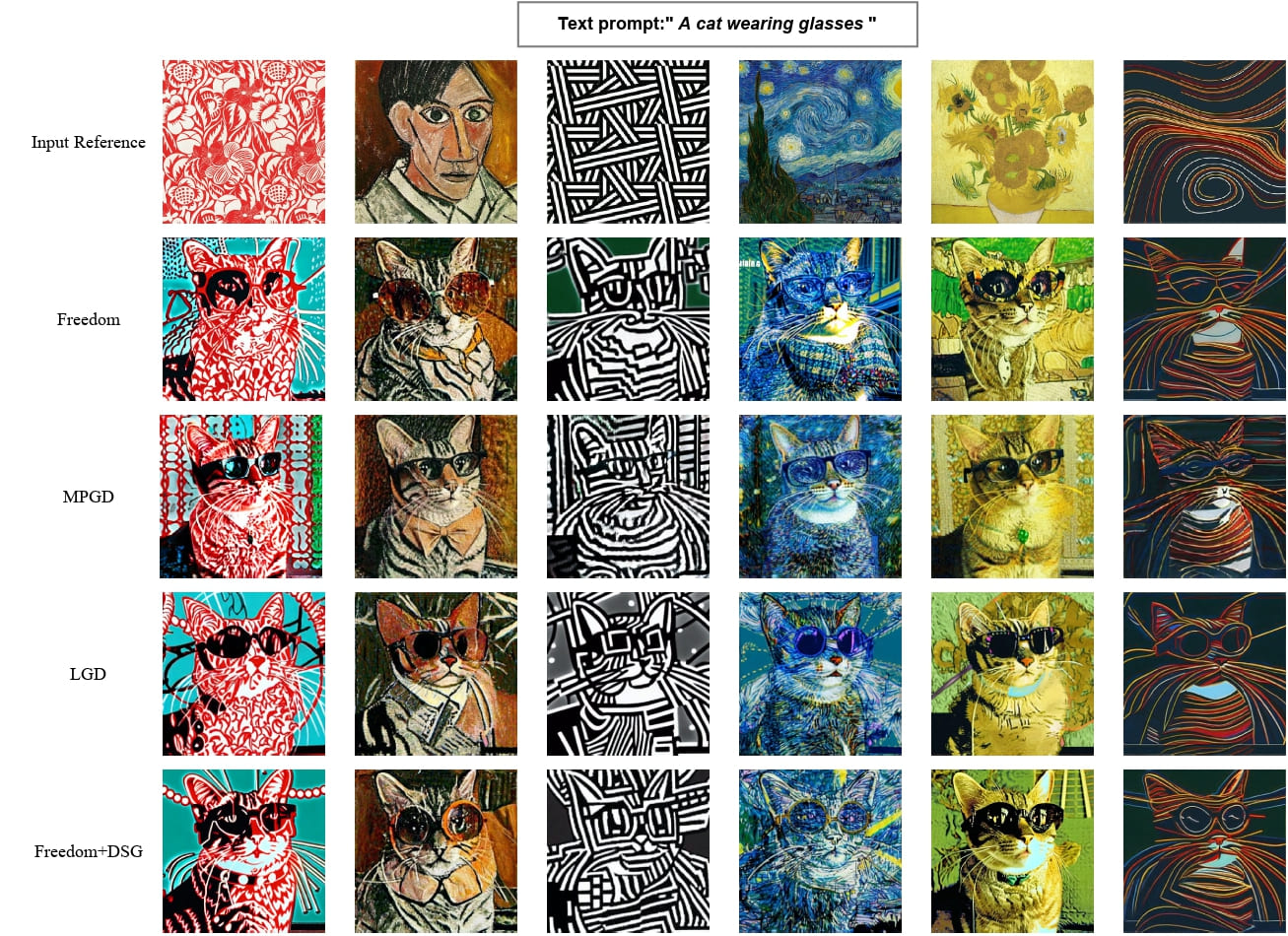}
  \caption{The qualitative results in Text-Style Guidance using Stable Diffusion.}
  \label{fig:text_style}
\end{figure*}

In Text-Style Guidance, we follow the experiment setting of Freedom \cite{freedom} and choose Freedom, LGD, and MPGD \cite{freedom, Lossguided, manifold} as the baseline methods for comparisons. For the hyperparameter setting, we follow the original setting of Freedom and MPGD. For LGD,  we set the number of Monte Carlo samples to 2, covariance to $0.1\sqrt{1-\alpha_t}$, and step size same as Freedom. The qualitative results are shown in Figure \ref{fig:text_style}.

\subsection{More Experimental Details and Results for the Inverse Problems in ImageNet}

\begin{figure*}[htbp]
  \centering
  \includegraphics[width=1\linewidth]{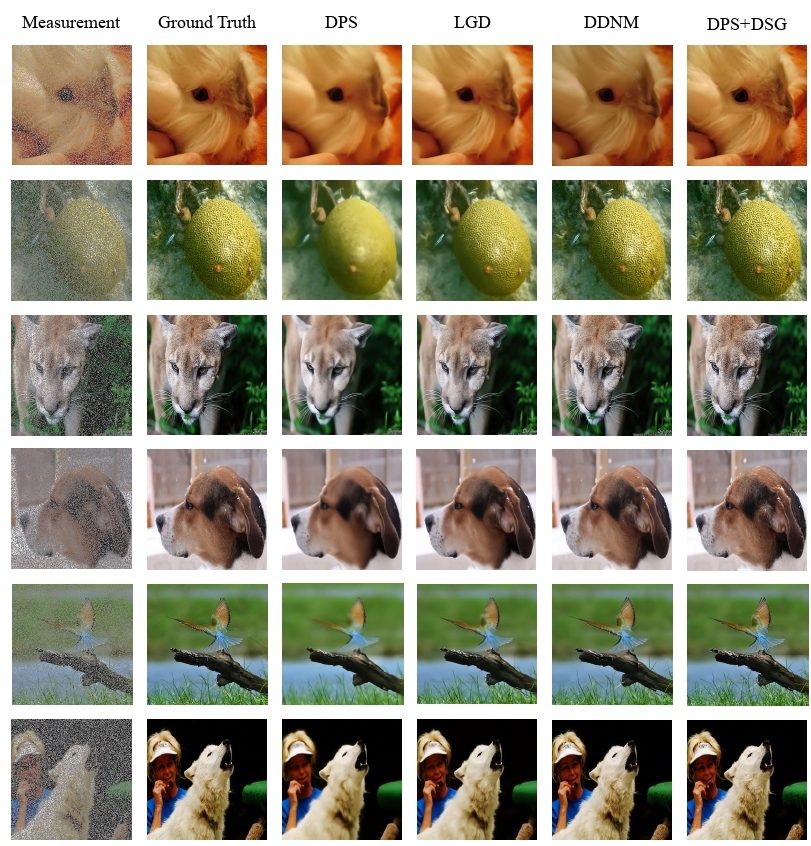}
  \caption{Additional qualitative results of Inpainting in Imagenet256*256.}
  \label{fig:imagenetInpainting}
\end{figure*}

\begin{figure*}[htbp]
  \centering
  \includegraphics[width=0.9\linewidth]{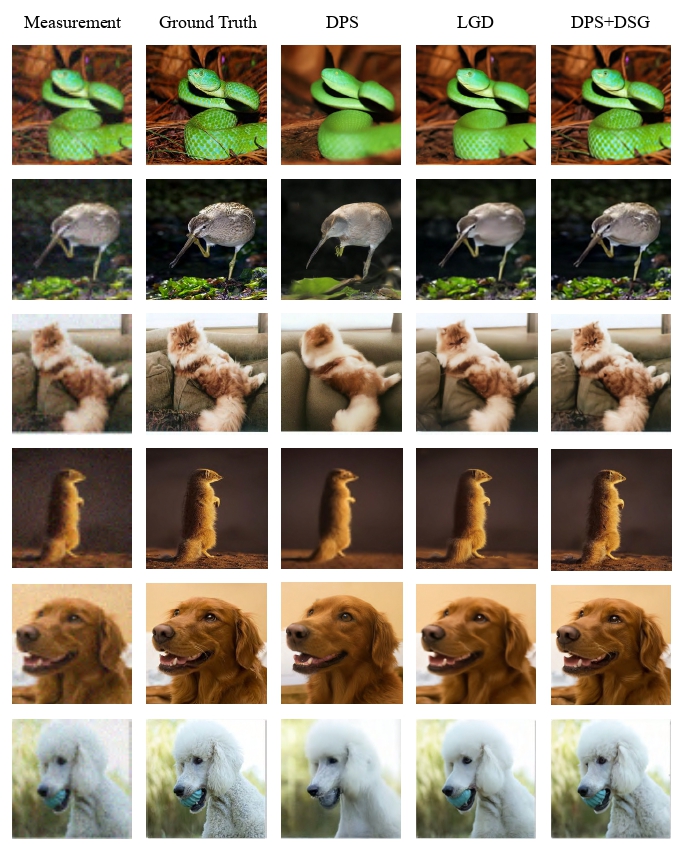}
  \caption{Additional qualitative results of Super-resolution in Imagenet256*256.}
  \label{fig:imagenetSR}
\end{figure*}

\begin{figure*}[htbp]
  \centering
  \includegraphics[width=0.9\linewidth]{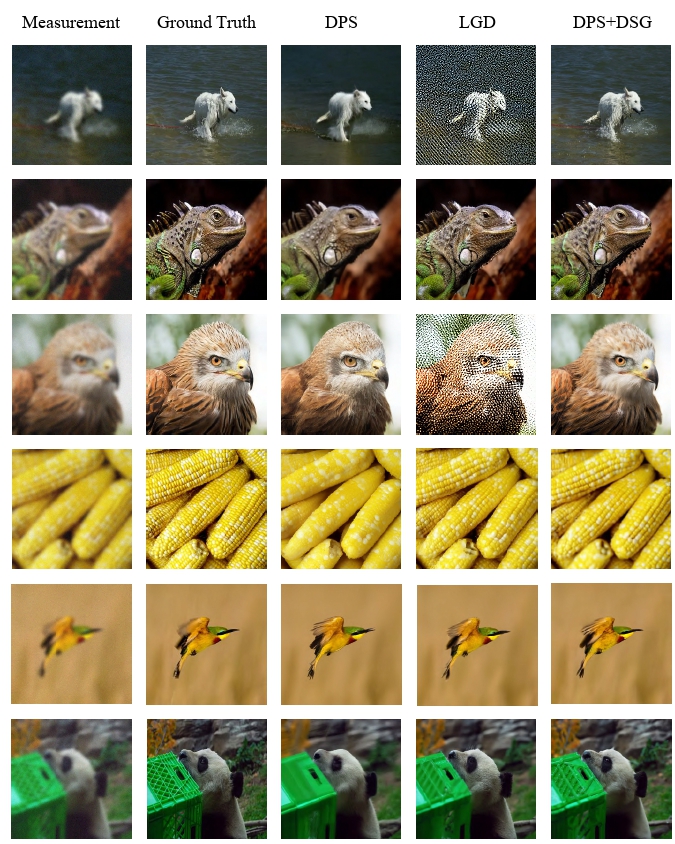}
  \caption{Additional qualitative results of Gaussian-deblurring in Imagenet256*256.}
  \label{fig:imagenetGD}
\end{figure*}

We validate the performance of DSG in inverse problems (Inpainting, Super-resolution, and Gaussian deblurring) in the 1k validation set of ImageNet256*256 using the same setting in Sec 5.1. We choose DPS, LGD, and DDNM as the baseline. For DPS and DDNM, we follow the hyperparameter setting of the original paper. For LGD, we set the number of Monte Carlo samples to 10, covariance to $\sigma_t / \sqrt{1+\sigma_t^2}$, and the same step size as DPS. As shown in Table \ref{tb:imagenet}, Figure \ref{fig:imagenetInpainting}, Figure \ref{fig:imagenetSR}, and Figure \ref{fig:imagenetGD}, our DSG outperforms other baselines in ImageNet256*256.

It's worth noting that DDNM \cite{ddnm} is \textbf{primarily applicable to linear inverse problems} and cannot be directly applied to nonlinear cases due to its direct access to the noise scale of measurement, the forward operator, and its pseudo-inverse. However, despite these differences, DSG still offers superior performance compared to DDNM.

\begin{table*}[h]
    \vspace{-0.5cm}
    \caption{Quantitative results in Linear Inverse Problem in ImageNet 256*256}
    \centering
    \footnotesize
    \tabcolsep=0.22cm
    \resizebox{\textwidth}{!}{
    \begin{tabular}{c |  ccc | ccc | ccc | c | }
        \toprule[1.2pt]
          \multirow{2}{*}{Methods}  &  \multicolumn{3}{c}{Inpainting} \vline & \multicolumn{3}{c}{Super resolution} \vline & \multicolumn{3}{c}{Gaussian deblurring}  \vline \\
        \rule{0pt}{8pt}
            & SSIM$\uparrow$ & PSNR$\uparrow$ & LPIPS↓ & SSIM$\uparrow$ & PSNR$\uparrow$ & LPIPS↓ & SSIM$\uparrow$ & PSNR$\uparrow$ & LPIPS↓ \\
        \toprule[1.2pt]
   
        DPS~\cite{dps} & 0.828 &  28.57 & 0.189 & 0.635 & 23.73 & 0.317 & 0.471 & 19.89 & 0.402 \\
        LGD~\cite{Lossguided} & 0.825 &  28.11 & 0.191 & 0.633 & 23.13 & 0.318 & 0.367 & 16.77 & 0.474 \\
        DDNM~\cite{ddnm} & 0.875 &  28.76 & 0.125 & - & - & - & - & - & - \\
        DPS+DSG(Ours) & \textbf{0.879} & \textbf{29.20} & \textbf{0.116} & \textbf{0.672} & \textbf{23.86} & \textbf{0.282} & \textbf{0.644} & \textbf{23.25} & \textbf{0.282}  \\

        \toprule[1.2pt]

    \end{tabular}
    }
    \vspace{-0.5cm}
    \label{tb:imagenet}
\end{table*}

\label{sec:app_imagenet}

\section{Additional Qualitative Results}

We provided additional qualitative results to demonstrate that DSG can plug in other training-free methods while improving their performance.
\begin{figure*}[htbp]
  \centering
  \includegraphics[width=1\linewidth]{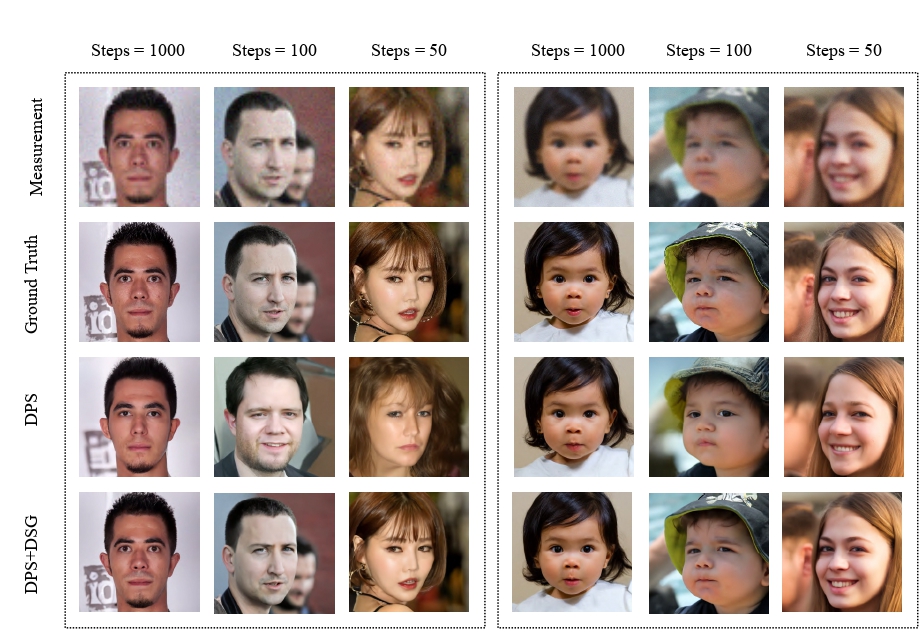}
  \caption{Extra qualitative examples of Super-resolution (left) and Gaussian deblurring (right) in the FFHQ dataset.}
\end{figure*}

\begin{figure*}[htbp]
  \centering
  \includegraphics[width=0.9\linewidth]{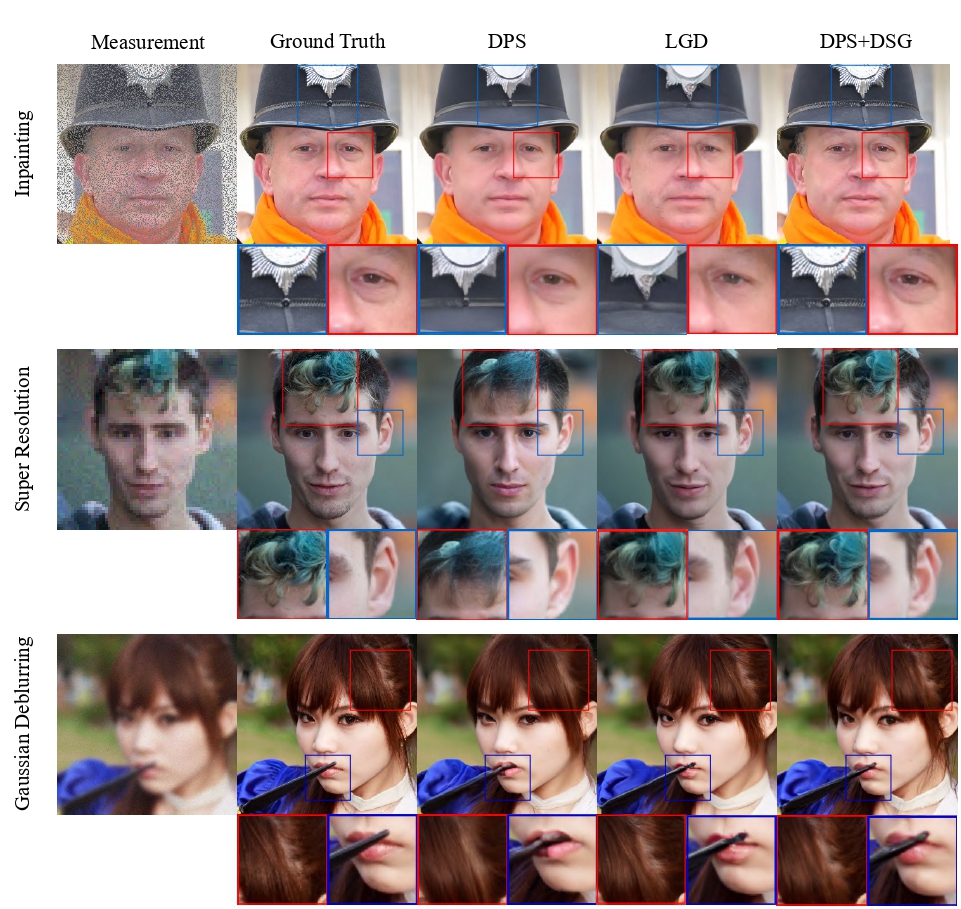}
  \caption{Enlargement of the qualitative results using DPS+DSG compared to DPS.}
\end{figure*}

\begin{figure*}[htbp]
  \centering

  \includegraphics[width=0.9\linewidth]{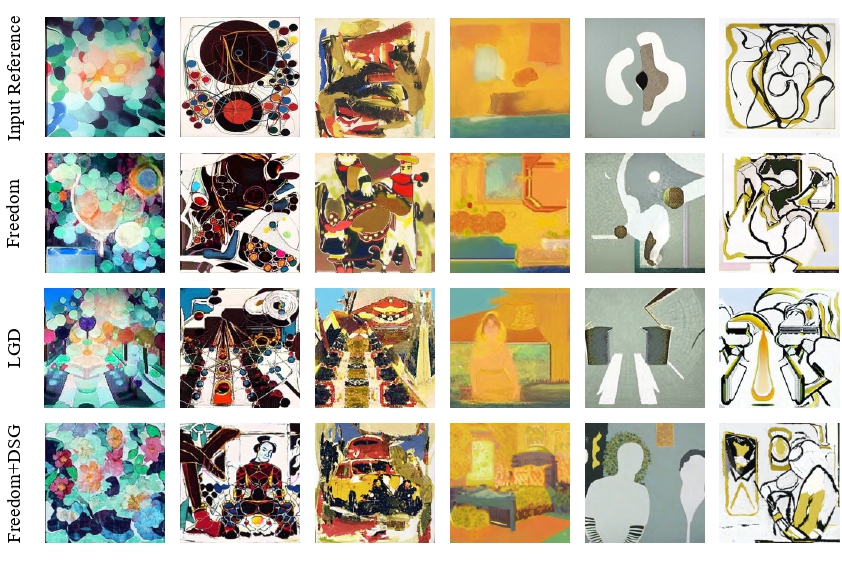}
  \caption{Additional qualitative results in Style Guidance using pre-trained Stable Diffusion.}
\end{figure*}

\begin{figure*}[htbp]
  \centering

  \includegraphics[width=0.9\linewidth]{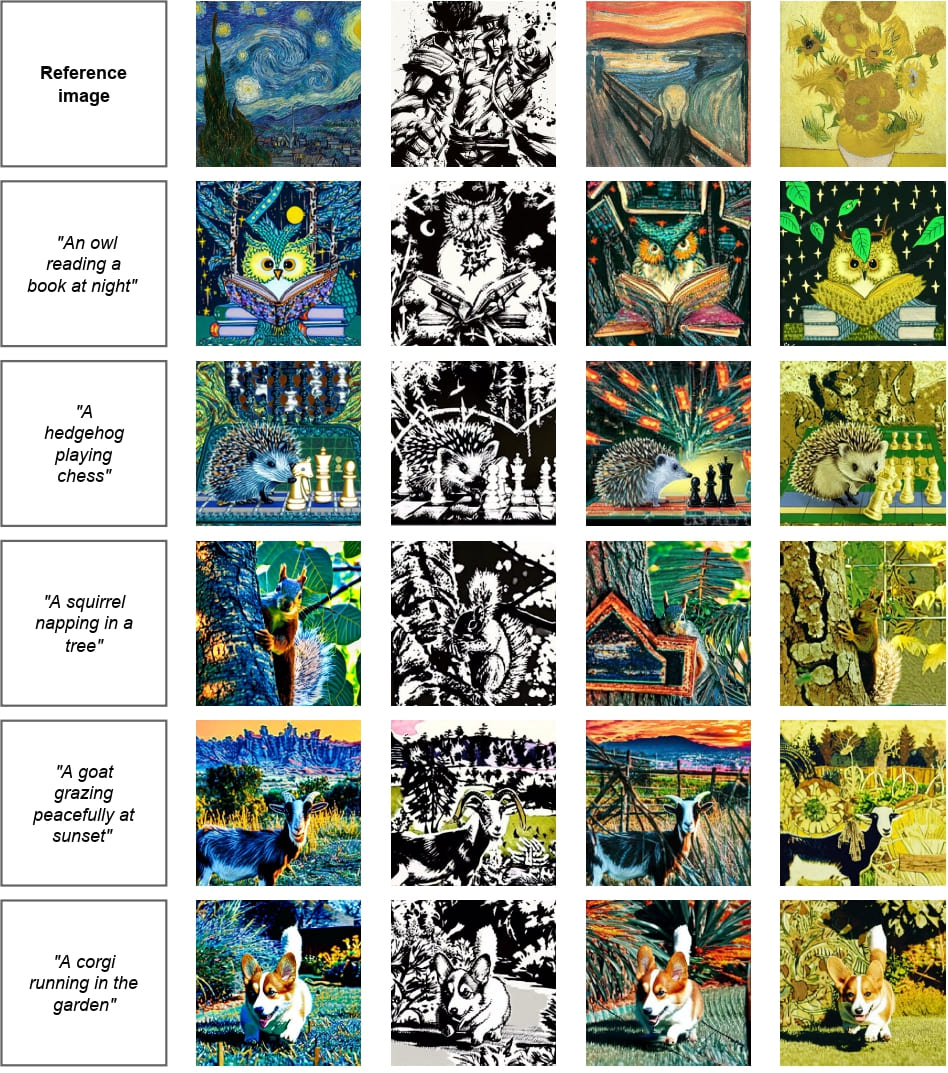}
  \caption{Additional qualitative results of DSG in Text-Style Guidance using pre-trained Stable Diffusion.}
\end{figure*}

\end{document}